\documentclass[letterpaper, 10 pt, conference]{ieeeconf}  
\usepackage{graphicx} 

\usepackage{booktabs}
\usepackage[pagebackref=true,breaklinks=true,letterpaper=true,hidelinks,bookmarks=false]{hyperref}

\usepackage[accsupp]{axessibility}  

\usepackage{graphicx}
\usepackage{booktabs}
\usepackage{threeparttable}
\usepackage{xspace}
\usepackage[dvipsnames]{xcolor}
\usepackage{comment}
\usepackage{adjustbox}
\usepackage{array}
\usepackage{makecell}
\usepackage{pifont}
\usepackage{multirow}
\usepackage{flushend}
\usepackage{colortbl}
\let\labelindent\relax
\usepackage{enumitem}
\usepackage{pgf}
\usepackage{import}
\usepackage{comment}
\usepackage[normalem]{ulem}
\usepackage{sidecap}
\newcolumntype{R}[2]{%
    >{\adjustbox{angle=#1,lap=1.3\width-(#2)}\bgroup}%
    l%
    <{\egroup}%
}
\usepackage[]{pgfplots,pgfplotstable}
\usepackage[]{tikz}
\usepackage{tabularx}

\makeatletter
\newcommand\myparagraph{\@startsection{paragraph}{4}{\z@}%
    {-6\p@ \@plus -3\p@ \@minus -3\p@}%
    {-0.5em \@plus -0.22em \@minus -0.1em}%
    {\normalfont\normalsize\itshape}}
\renewcommand\paragraph{\@startsection{paragraph}{4}{\z@}%
    {1.25ex \@plus 1ex \@minus .2ex}%
    {-1em}%
    {\normalfont \normalsize \bfseries}}
\makeatother

\definecolor{myblue}{rgb}{0.19, 0.55, 0.91}
\definecolor{myred}{rgb}{0.82, 0.1, 0.26}
\definecolor{MyGreen}{RGB}{0, 104, 55} 
\definecolor{MyRed}{RGB}{248, 3, 7} 
\definecolor{MyYellow}{RGB}{180, 180, 0} 
\newcommand{\cmark}{{\textcolor{MyGreen}{\ding{51}}}}%
\newcommand{\xmark}{{\textcolor{MyRed}{\ding{55}}}}%
\newcommand{\NS}[1]{\textcolor{Black}{#1}}

\newcommand{\SG}[1]{\textcolor{Black}{#1}}

\def\ours{\texttt{\textcolor{red}{R3D}PA}\xspace} 

\newcommand{\cropcenter}[2]{%
  \includegraphics[width=#1,clip,trim=160 0 160 0]{#2}
}

\definecolor{brightpink}{rgb}{1.0, 0.0, 0.5}

\IEEEoverridecommandlockouts                              

\title{\LARGE \bf

\ours: Leveraging 3D Representation Alignment and RGB Pretrained Priors for LiDAR Scene Generation

}

\author{Anonymous Submission
}

\author{Nicolas Sereyjol-Garros$^\dagger$, Ellington Kirby, Victor Besnier$^\ddagger$, Nermin Samet$^\ddagger$ 
\thanks{All authors are affiliated with Valeo.ai, Paris, France. Email: \texttt{\{nicolas.sereyjol-garros, ellington.kirby, victor.besnier, nermin.samet\}@valeo.com}
}
\thanks{$^\dagger$ Corresponding author.}
\thanks{$^\ddagger$ Equal contribution for senior authorship.}%
}

\begin{document}

\maketitle

\begin{abstract}
LiDAR scene synthesis is an emerging solution to scarcity in 3D data for robotic tasks such as autonomous driving.
Recent approaches employ diffusion or flow matching models to generate realistic scenes, but 3D data remains limited compared to RGB datasets with millions of samples.
We introduce \ours, the first LiDAR scene generation method to unlock image-pretrained priors for LiDAR point clouds, and leverage self-supervised 3D representations for state-of-the-art results.
Specifically, we
(i) align intermediate features of our generative model with self-supervised 3D features, which substantially improves generation quality;
(ii) transfer knowledge from large-scale image-pretrained generative models to LiDAR generation, mitigating limited LiDAR datasets;
and (iii) enable point cloud control at inference for object inpainting and scene mixing with solely an unconditional model.
On the KITTI-360 benchmark \ours achieves state of the art performance.
Code and pretrained models are available at \url{https://github.com/valeoai/R3DPA}.

\end{abstract}    
 \section{Introduction}
\label{sec:intro}

Autonomous driving relies on the ability of vehicles to navigate complex environments safely and efficiently. Among various sensors such as cameras, GPS, or radar, LiDAR stands out as a key sensor due to its precise 3D measurements of vehicles, pedestrians, and drivable areas. 
However, collecting and annotating real-world  large-scale LiDAR point cloud datasets is notoriously expensive and time consuming, limiting the development of scalable autonomous driving systems. The challenge grows even more when accounting for variations in LiDAR sensor density, as each variation requires its own extensive dataset.

\begin{figure}[!htb]
    \centering
    \begingroup
    \footnotesize
    \def\svgwidth{\columnwidth} 
    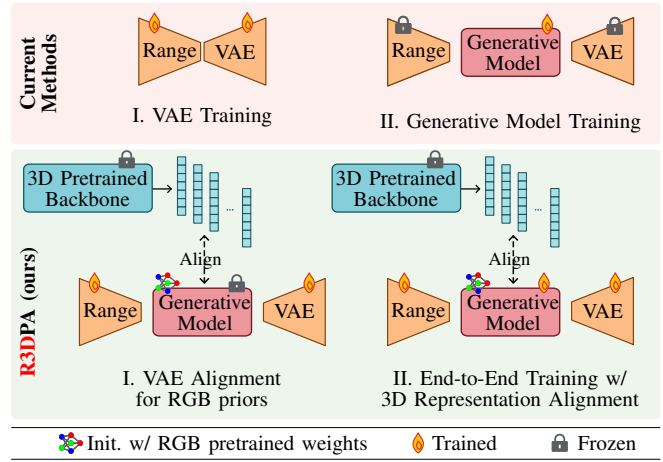
    \endgroup
    \caption{\textbf{\ours pipeline overview.} LiDAR point cloud generation from range images commonly follows a two-stage approach: the VAE is trained independently and then frozen, while the generative model is trained on its latent space. In contrast, our method leverages priors from a backbone pretrained on large-scale image datasets. \textbf{The alignment step trains the VAE from scratch} while initializing and freezing the generative model with pretrained weights. 
    This stage ensures that the latent space of our newly trained VAE remains compatible with the knowledge of the pretrained generative model.
    \textbf{We then jointly optimize the VAE encoder and the generative model under the supervision of 3D representations}. Range VAE denotes a model trained on range images.}
    \label{fig:tuning}
\end{figure}

These challenges have motivated research into generating synthetic LiDAR datasets, including LiDAR simulation in virtual environments \cite{gschwandtner2011blensor, fang2020augmentedLiDAR, fang2021LiDARaug, yue2018LiDARpcgen}, as well as generative learning approaches \cite{zyrianov2022learning} \cite{xiong_ultralidar_2023}. More recently, diffusion models have demonstrated remarkable success in capturing complex data distributions for image synthesis, positioning them as a promising direction for LiDAR data generation to further bridge the realism gap.

Early diffusion models for image generation relied on pixel-level space \cite{song2019generative, ho_denoising_2020, song_score-based_2020}. To address the challenges posed by the high dimensionality of pixel-level representations, latent diffusion models were later introduced, offering both faster training and higher-quality generation \cite{rombach_high-resolution_2022, podell2024sdxl, esser2024scaling}. 
More recently, RGB image synthesis has been further advanced through representation alignment~\cite{yu_representation_2025, leng_repa-e_2025, tian_u-repa_2025, yao_reconstruction_2025}, where diffusion models are trained to align their internal representations with powerful self-supervised features such as those from DINOv2~\cite{oquab2024dinov2}. These methods show that aligning with pretrained representations accelerates training and improves generation quality.

3D LiDAR point cloud scene generation has followed a similar path. Some diffusion models target point-level generation \cite{nunes2024scaling, martyniuk2025lidpm, xiong_ultralidar_2023},  while other approaches focus on range image representations of point clouds~\cite{zyrianov2022learning,nakashima_fast_2025,nakashima_lidar_2024}, which offer structural advantages over unorganized point sets and enable more efficient use of latent space \cite{hu_rangeldm_2024, ran_towards_2024}.

However, recent advancements in state-of-the-art image diffusion models have not been leveraged for 3D data. First, existing approaches for LiDAR scene generation do not exploit powerful self-supervised 3D representations similar to REPA \cite{yu_representation_2025}. 
Second, they are trained entirely from scratch, usually on mid-scale datasets such as nuScenes \cite{caesar2020nuscenes} or KITTI-360 \cite{liao2022kitti}, whereas image diffusion and flow matching models typically leverage large-scale datasets such as ImageNet \cite{deng2009imagenet} or LAION \cite{schuhmann2022laion}.

Inspired by recent advances in representation alignment for image generation, we aim to enhance representation learning in LiDAR scene generation from range images. We argue that such representations are essential for achieving realistic LiDAR scene generation, enabling tasks such as completion and denoising. 
We propose \ours, the first unconditional LiDAR point cloud generator to leverage both RGB image priors and 3D representations for high-fidelity LiDAR scene generation. \NS{We provide an overview of our method in Figure \ref{fig:tuning}.} Our contributions:
\begin{itemize}
    \item To our knowledge, \ours is the first work to repurpose the weights of a pretrained natural image flow matching (FM) model for LiDAR scene synthesis. While directly initializing a 3D generative model with these weights does not improve generation quality, our proposed training strategy enables the transfer of rich priors learned from large-scale RGB image collections, compensating for the limited scale of LiDAR datasets.    
    \item We are also the first to leverage self-supervised 3D features for representation alignment in LiDAR point cloud generation. We demonstrate that, similar to image generation, representation alignment with self-supervised 3D features significantly improves generation quality.
    \item Finally, as our model is aligned with self-supervised 3D features during training, it enables controllable scene editing at inference for free. By incorporating feature guidance into our unconditional model, we demonstrate two applications: object inpainting and scene mixing.
\end{itemize}
On the standard KITTI-360 benchmark, our experiments show that our method significantly improves generation quality of \NS{3D LiDAR point clouds}, surpassing the previous state of the art by at least 17\%.

\section{Related works}
\label{sec:rw}

\paragraph{LiDAR Scene Generation}

Some LiDAR scan generation methods, such as UltraLiDAR~\cite{xiong_ultralidar_2023} and OLiDM~\cite{yan2025olidm}, directly take point clouds as input.
To control for non-uniform and large point cloud sizes, these methods rely on encoding and pre-processing strategies such as voxelization or padding to create uniform point clouds \cite{kirby2024logen}.
The second and most widely used modality, range images, requires no costly pre-processing, and builds on top of existing classic image architectures like UNet~\cite{milioto_rangenet_2019} or ViT~\cite{ando_rangevit_2023}.
Range images enable an easier fusion between modalities and the possibility to use large pretrained models~\cite{buburuzan_mobi_2025}.

First attempts to use diffusion on range images opted for pixel space diffusion~\cite{zyrianov2022learning, nakashima_lidar_2024,zhu2025spiral}, which comes at the price of a costly generation process limiting the size of the network. More recent works, such as~\cite{ran_towards_2024,hu_rangeldm_2024}, switch to a latent diffusion approach~\cite{rombach_high-resolution_2022}.
Conversely, R2Flow~\cite{nakashima_fast_2025} argues that diffusion in pixel space is essential for high-fidelity results. To reduce inference time, it introduces a rectified flow model that lowers the number of required sampling steps. In contrast, we argue that latent diffusion can also deliver high-quality generation, provided the variational autoencoder is trained with an adversarial loss as in \cite{hu_rangeldm_2024}.

Significant advancements in range image generation architectures include the introduction of circular padding for convolutions in the UNet~\cite{milioto_rangenet_2019,ran_towards_2024,nakashima_lidar_2024}, the use of autoencoders as in LiDM~\cite{ran_towards_2024}, and improvements in range image projection~\cite{triess_scan-based_2020, hu_rangeldm_2024}.
\cite{triess_scan-based_2020} points out that the assumption of translation equivariance in convolutions is not fully satisfied for range images, due to distance dilation as a function of pitch.
LiDARGen~\cite{zyrianov2022learning} and R2DM~\cite{nakashima_lidar_2024} address translation equivariance by introducing spatial Fourier features of the angular coordinates as an additional input channel to a UNet. These features act as positional encodings to generate front view centered range images.
In contrast, we employ a transformer architecture, which naturally incorporates positional information without requiring such encodings.

\paragraph{Vision Foundation Models for Generation} 
Image synthesis has progressed from GANs \cite{goodfellow2014generative}, to autoregressive \cite{van2016conditional}, diffusion \cite{ho_denoising_2020}, and flow matching (FM) models \cite{lipmanflow}. 
While GANs enable fast generation, they often suffer from limited coverage and training instability \cite{brock2018large}.
On the other hand, diffusion methods offer stable training with simple objectives but rely on slow, iterative sampling \cite{song_score-based_2020}.
Recent hybrid approaches, such as \cite{rombach_high-resolution_2022}, address these trade-offs leveraging VAEs (variational autoencoders) and GANs to compress images into a smaller latent space, and a diffusion model for synthesis, improving both efficiency and fidelity.

Pioneering work REPA \cite{yu_representation_2025} observes that training speed in latent diffusion models is mainly limited by learning a meaningful internal representation of images. To address this, it guides the learning process by aligning intermediate projections with the feature maps of the DINOv2 encoder \cite{oquab2024dinov2}, which results in faster training and improved generation quality.
Building on this idea, subsequent works have examined the effectiveness of this additional loss \cite{wang_repa_2025} and its application to UNet architectures \cite{tian_u-repa_2025}.
\cite{yao_reconstruction_2025} improves reconstruction and generation quality by introducing a latent alignment loss that, unlike REPA, directly operates in the VAE latent space before diffusion training.
The REPA-E~\cite{leng_repa-e_2025} training method shows that the REPA loss can serve as a useful signal for the encoder, allowing end-to-end training. By backpropagating the REPA loss both to the diffusion model and the VAE, it jointly guides their internal representations, leading to improved alignment and generation quality.

\paragraph{Self-supervised Pretraining of LiDAR Scenes}
Self-supervised learning methods on images have provided strong general backbones yielding representations useful for a wide variety of downstream tasks \cite{oquab2024dinov2}. 
Recent work has aimed to develop analogous backbones for 3D LiDAR point clouds as foundation models.
SONATA \cite{wu2025sonata} follows a self-distillation pretraining strategy, using only the encoder of a UNet as the backbone.
ScaLR \cite{puy2024three} instead distills DINOv2~\cite{oquab2024dinov2} features into its 3D backbone using paired LiDAR point clouds and camera images, aligning 3D features with those extracted by the pretrained 2D backbone.
Along the same line, OpenScene~\cite{peng2023openscene} distills CLIP features into 3D features. 

\section{Method}
\label{sec:method}

We aim to enhance 3D LiDAR scene generation by combining rich 3D representations with pretrained RGB image priors learned from large-scale image datasets. 

\NS{
To exploit the knowledge learned from large-scale RGB image dataset, we initialize our latent FM model with weights pretrained on such data.
} 
\begin{figure}[t!]
\centering
\def\svgwidth{0.99\linewidth} 
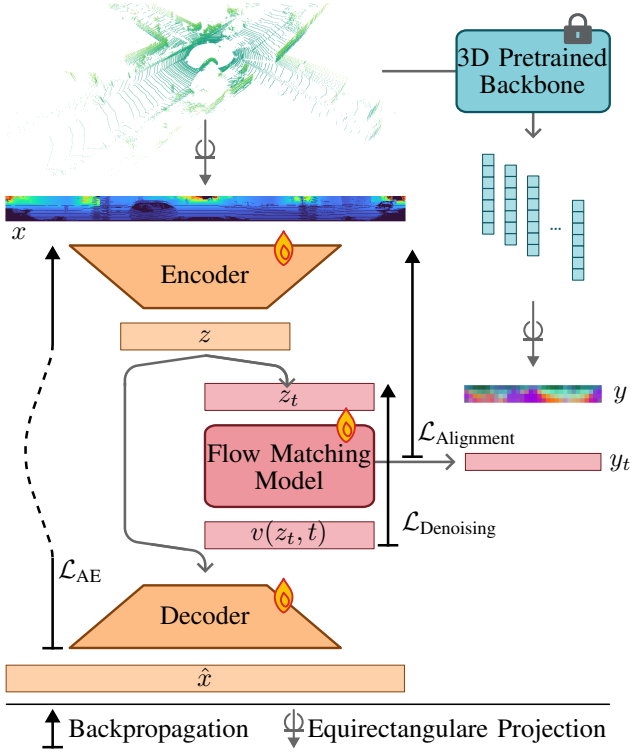
\caption{\textbf{End-to-end training.} We jointly train the VAE and the flow matching (FM) model with the additional alignment loss $\mathcal{L}_{\textrm{Alignment}}$, which aligns the internal representations of the FM model with 3D point features $y$. At each step, the VAE is updated with $\mathcal{L}_{\textrm{AE}} + \lambda_1 \mathcal{L}_{\textrm{Alignment}}$, followed by updating the FM model with $\mathcal{L}_{\textrm{Denoising}} + \lambda_2 \mathcal{L}_{\textrm{Alignment}}$.
$z_t$ is the interpolated noisy latent vector derived from the clean latent $z$.
The input point cloud and its associated features are projected onto the range image $x$ and the feature map $y$, respectively, using an equirectangular projection.
}
\label{fig:pipeline}
\end{figure}
However, directly fine-tuning from \NS{pretrained} weights causes a collapse of prior knowledge due to the domain gap between natural-image latents and range-image latents.
To address this issue, we introduce a VAE alignment training recipe, which is guided by 3D representation alignment for cross-modality adaptation.
We call this unified framework \ours, which operates on range images while leveraging both 2D priors from natural images and 3D representation alignment. 
An overview of our framework is shown in Figure~\ref{fig:tuning}.

In \ours, 3D representations from a 3D pretrained backbone are used to compute an alignment loss that supervises the internal activations of the FM model. The first stage, VAE alignment, freezes the initialized FM model while training the VAE from scratch. Guided by the alignment loss, the VAE adapts its latent space to the input space of the FM model. In the second stage, end-to-end training, we apply joint alignment: the alignment loss propagates gradients to both the VAE and the FM model, ensuring that the autoencoder also benefits from a FM model signal rather than leaving the adaptation solely to the FM model. We illustrate our end-to-end training in Figure~\ref{fig:pipeline}.

For the alignment loss, each LiDAR point cloud is encoded with a pretrained 3D feature extractor to obtain point-level self-supervised features. An equirectangular projection then maps the unstructured point cloud into a 2D feature grid that is geometrically aligned with the FM model’s internal representations. This process distills knowledge from the pretrained 3D foundation model and establishes a shared, expressive latent space between the VAE and the FM model.

\subsection{Preliminaries}
This section provides the necessary background for our method. Following prior work \cite{rombach_high-resolution_2022,hu_rangeldm_2024,ran_towards_2024}, we generate LiDAR point clouds by applying FM in the latent space of a VAE.

\textbf{Equirectangular Projection.}
We represent unstructured point clouds as range images where pixel coordinates come from discretized yaw and pitch, pixel values store depth, and only the closest point is kept per pixel.
This representation, while not being lossless, allows a much faster processing. 
More precisely, the pixel coordinate  $(i,j)$ of a point $(p_x,p_y,p_z)$  projected onto a grid of size $H$x$W$ are
\begin{equation}
\begin{pmatrix}
i \\
j
\end{pmatrix}
=
\begin{pmatrix}
\left( 1 - \left( \arcsin(p_z/ r) + f_{\text{down}} \right) f_v^{-1} \right) H
\\
\frac{1}{2}\left(1 - \arctan(p_y/p_x) \pi^{-1}\right) W
\end{pmatrix}
\label{eq:proj}
\end{equation}
where $f_v = |f_{up} - f_{down}|$ is the vertical field of view of the LiDAR sensor and $r=\sqrt{p_x^2+p_y^2+p_z^2}$ is the depth.

\textbf{VAE Training.}
We adopt the VAE architecture of~\cite{ran_towards_2024}, which uses horizontal kernels \NS{and circular padding} in convolution and downsampling layers to account for the wide aspect ratio of LiDAR range images.  
\NS{We encode range image $x$ with encoder $\mathcal{E}$ to a compact latent vector $z=\mathcal{E}
(x)$} with downsampled spatial dimensions and an upsampled channel dimension. Overall, the smaller dimension of the latent space makes the denoising process faster. 
We decode with $\mathcal{D}$ latent vector z to produce $\hat{x} = \mathcal{D}(z)=\mathcal{D}(\mathcal{E}(x))$ a reconstructed range image.

The overall VAE objective is $\mathcal{L}_{\text{AE}} = \mathcal{L}_{\text{MSE}} + \mathcal{L}_{\text{KL}} + \mathcal{L}_{\text{GAN}}$ similar to~\cite{rombach_high-resolution_2022}.  The reconstruction loss $\mathcal{L}_{\text{MSE}}$ is computed between the input range image $x$ and its reconstruction $\hat{x}$. The term $\mathcal{L}_{\text{KL}}$ enforces the encoder's latent distribution to remain close to a Gaussian prior, ensuring a smooth and sampling-friendly latent space. To mitigate the typical blurriness of VAE outputs, we also include an adversarial loss $\mathcal{L}_{\text{GAN}}$, which uses a patch-wise discriminator to encourage realistic and sharp reconstructions, as motivated by \cite{nakashima_fast_2025}. The discriminator is implemented as a convolutional network and trained jointly with the VAE.

\textbf{LiDAR Scene Generation with Flow Matching.}
We generate LiDAR scenes by producing latent representations with a stochastic interpolant~\cite{ma_sit_2024}, which are subsequently decoded into range images. These range images are then unprojected to obtain point clouds.

Stochastic interpolants unify flow- and diffusion-based generative models~\cite{albergo2023stochastic}. In our work, we adopt a simple linear interpolation process that gradually transforms noise $\varepsilon\sim\mathcal{N}(\mathbf{0},\mathbf{I})$ into latent data $\mathbf{z}_*\sim p(z)$ in reverse time, defined as $z_t = (1-t)\mathbf{z}_*+t\varepsilon$ with $ t\in[0,1]$. The target $\mathbf{z}_*$ can then be generated either deterministically or stochastically, using an ordinary or a stochastic differential equation, respectively. In both cases, only a single estimator of the velocity field is required, given by $ \mathbf{v}(z,t) = \mathbb{E}[\varepsilon-\mathbf{z}_*|z_t=z]$. We optimize our FM network $v$ to be an estimator of $\mathbf{v}$ with the following objective $\mathcal{L}_{\text{Denoising}} = \mathbb{E}[\|v(z_t,t)-(\varepsilon - \mathbf{z}_*)\|^2]$
We parametrize the estimator of the vector field using a transformer architecture. 
 
\subsection{R3DPA}

\begin{figure}[t!]
    \centering
    \begin{minipage}{0.9\columnwidth}
        \centering
        \includegraphics[width=\linewidth]{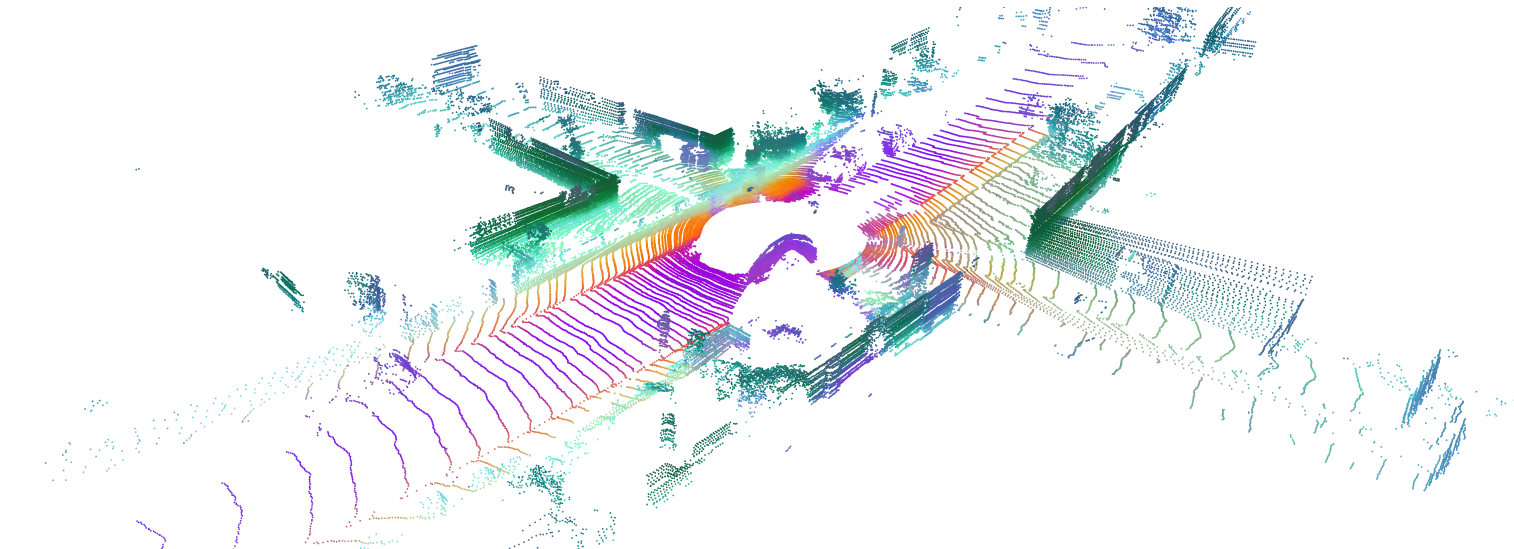}
        \label{fig:mask}
    \end{minipage}
    \hfill
    \begin{minipage}{0.9\columnwidth}
        \centering
        \includegraphics[width=\linewidth]{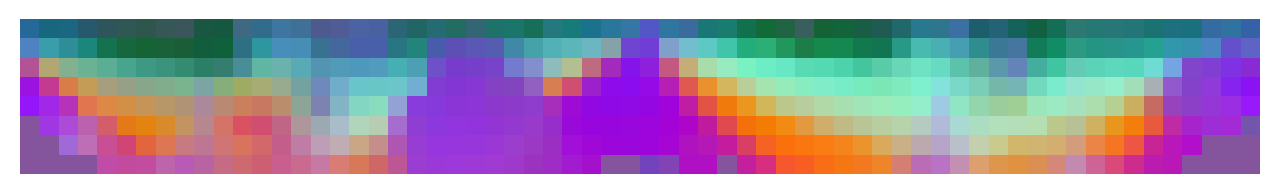}
    \end{minipage}
    \caption{\textbf{PCA visualization of 3D features and their projection.} The 768-dimensional 3D features \cite{puy2024three} are reduced to three principal components, which are then mapped to RGB colors for visualization.
    (Top) Point features in LiDAR scene, (Bottom) Projected features at grid size 8$\times$64, matching the internal representation of the transformer \cite{ma_sit_2024}.}
    \label{fig:scalr_features}
\end{figure}

\textbf{3D Representation Alignment.}
The goal of representation alignment is to supervise the internal representations of FM transformers using features extracted from a pretrained encoder. 
Following~\cite{wang_repa_2025, leng_repa-e_2025}, we first project the transformer output from the middle layer into a feature map of size $H \times W$ using a projection layer. 
We then extract 3D point features from the corresponding LiDAR scene using a pretrained 3D backbone. However, these 3D point features are not directly usable due to their large number and unstructured nature.  
We therefore project them onto a grid using an equirectangular projection (\ref{eq:proj}), with the grid size matching the spatial dimensions of projection layer output.
Points falling into the same grid cell are averaged, resulting in a feature grid $y \in \mathbb{R}^{H \times W \times D}$ where $D$ is the dimension of 3D features. 
Figure \ref{fig:scalr_features} shows a PCA visualization of projected features. 
Since each grid cell can be mapped one-to-one to a FM token, the alignment loss can be applied as follows: 

\begin{equation}
\mathcal{L}_{\textrm{Alignment}}(\theta,\phi) = -\mathbb{E} \Bigg[ \frac{1}{H} \frac{1}{W} \sum_{j=1}^H \sum_{i=1}^W \langle y_{(i,j)}, y_{(i,j)}^t \rangle \Bigg], 
\label{eq:align}
\end{equation}

Here, $y \in \mathbb{R}^{H \times W \times D}$ denotes the normalized and projected 3D features, 
while $y^t$ is the normalized  and projected internal representation of the model from noisy latent at timestep $t$,  
$\theta$ is the parameters of the FM model and $\phi$ is the parameters of the projection head.

 \begin{figure*}[t!]
    \centering
    \begin{tabular}{ccc}
        \begin{minipage}{0.49\textwidth}
            \vspace{+0.25em}
            \centering
            \includegraphics[width=\linewidth]{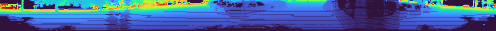} \\[0.5ex]
            \includegraphics[width=\linewidth]{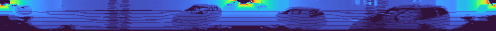} \\[0.5ex]
            \includegraphics[width=\linewidth]{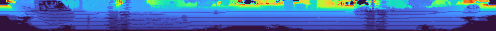} \\[0.5ex]
            \includegraphics[width=\linewidth]{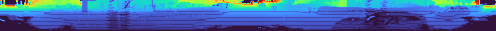} \\[0.5ex]
            \small \textrm{Range Images}
        \end{minipage}
        \begin{minipage}{0.24\textwidth}
            \centering
            \includegraphics[width=\linewidth]{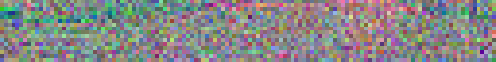} \\[0.5ex]
            \includegraphics[width=\linewidth]{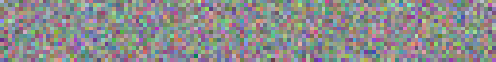} \\[0.5ex]
            \includegraphics[width=\linewidth]{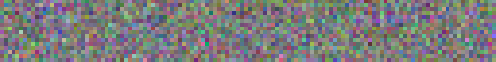} \\[0.5ex]
            \includegraphics[width=\linewidth]{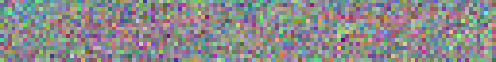} \\[0.5ex]
            \small \textrm{Standard VAE latents}
        \end{minipage}
        \begin{minipage}{0.24\textwidth}
            \centering
            \includegraphics[width=\linewidth]{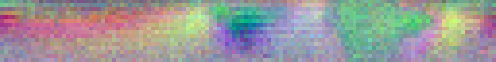} \\[0.5ex]
            \includegraphics[width=\linewidth]{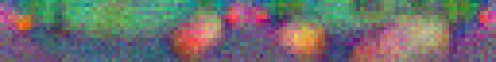} \\[0.5ex]
            \includegraphics[width=\linewidth]{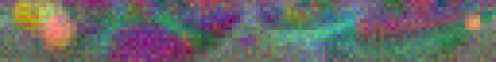} \\[0.5ex]
            \includegraphics[width=\linewidth]{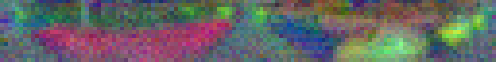} \\[0.5ex]
            \small \textrm{End-to-end VAE latents}
        \end{minipage} \\
    \end{tabular}
    \caption{\textbf{Effect of end-to-end training on latents.} (Left) Input range images. (Middle) Corresponding PCA visualizations of latents obtained using the standard VAE paradigm, such as \cite{ran_towards_2024}, where the VAE is trained independently on range images. (Right) PCA visualizations of latents from our end-to-end training with 3D representation alignment. Latents obtained through end-to-end training are more expressive, allowing a clear distinction of objects and background in the scene.}
    \label{fig:latent_visu}
\vspace{-0.5cm}
\end{figure*}

\textbf{VAE Alignment for RGB Image Priors.}
Our VAE alignment step enables us to exploit natural image pretrained FM model weights by transferring their knowledge to range images, which are limited in dataset size.  
To this end, we initialize our model with the weights of a FM model trained on the same backbone using a significantly larger dataset of natural images. 
However, we observe that directly training with these weights yields no performance gain (see Table \ref{tab:abl_vae}), which is expected given the domain gap between natural-image latents and range-image latents. 
To address this, we introduce VAE alignment: the FM model is initialized with RGB image pretrained weights, with the exception of the feature projection head and the first layer of the transformer patch embedding, which is trained to resolve the latent channel mismatch, while the rest of the model is kept frozen and only the VAE is trained. 
\SG{During training, we include the alignment loss (\ref{eq:align}) to the VAE training and use it as a guidance signal from the generative model to the encoder. This signal enforces the VAE to align with the internal representations of the pretrained FM backbone.} Once this step is complete, we unfreeze the FM model and proceed with full end-to-end training.

\textbf{End-to-end training.} End-to-end training jointly optimizes the VAE and the FM model (see Figure~\ref{fig:pipeline}).
During joint optimization, the VAE and SiT networks are updated with their respective objectives, both combined with $\mathcal{L}_{\text{Alignment}}$. Concretely, we optimize at each step (i) the VAE with $\mathcal{L}_{\text{AE}} + \lambda_1 \mathcal{L}_{\text{Alignment}}$, where $\mathcal{L}_{\text{Alignment}}$ updates only the encoder weights of VAE, and (ii) the FM model with $\mathcal{L}_{\text{Denoising}} + \lambda_2 \mathcal{L}_{\text{Alignment}}$.

FM models are typically trained in a space normalized to unit variance. Since the VAE latent space evolves during training, we follow~\cite{leng_repa-e_2025} and normalize the transformer input with a batch normalization layer. At inference, we fix the normalization using statistics accumulated with an exponential moving average.

End-to-end training improves alignment between the encoder and the FM model, leading to a more expressive latent space (see Figure~\ref{fig:latent_visu}) and higher generation quality.

\section{Experiments}
\label{sec:experiments}

\subsection{Experimental Setup}
\textbf{Datasets.} We conduct our state-of-the-art comparisons and ablation experiments on the KITTI-360 dataset \cite{liao2022kitti}. 
KITTI-360 contains images and LiDAR scans collected with a 64-beam LiDAR sensor across suburban areas of Karlsruhe, Germany, comprising approximately 80k scans from 9 sequences. 
For LiDAR scene generation, we follow the standard dataset split used in prior works \cite{zyrianov2022learning, nakashima_lidar_2024, nakashima_fast_2025}.  

To pretrain the flow matching (FM) model, we use ImageNet-1k \cite{liao2022kitti}, which contains over 1 million images.

\textbf{Implementation Details.}
We use 3D self-supervised features from ScaLR \cite{puy2024three}, which ranks among the state-of-the-art self-supervised backbones for LiDAR.
We implement our FM model using the base Scalable Interpolant Transformer~\cite{ma_sit_2024} with a patch size of 2 (SiT-B/2).
We train SiT-B/2 for 1M steps with a batch size of 256 using the Stable Diffusion VAE on ImageNet, following the official code and protocol of \cite{leng_repa-e_2025}.
During the VAE alignment stage, we train our VAE from scratch for 800k steps with a batch size of 32. After alignment, we perform end-to-end training for 1M steps, also with a batch size of 32. 
We set $\lambda_1$ and $\lambda_2$ to 1.5 and 0.5, respectively, as in \cite{wang_repa_2025}.
 
\textbf{Metrics.}  We evaluate the distance between the distributions of real and generated scenes using the Fr\'echet distance computed on different feature representations. 
Following \cite{ran_towards_2024}, we report Fr\'echet Range Image Distance (FRID), Fr\'echet Sparse Volume Distance (FSVD), and Fr\'echet Point Volume Distance (FPVD). 
In addition, we introduce Fréchet Localization Distance (FLD) to assess the quality of range images. 

LiDM \cite{ran_towards_2024} computes FRID using RangeNet++ \cite{milioto_rangenet_2019} trained on SemanticKITTI \cite{behley2019iccv}. 
However, we observed that FRID scores using RangeNet++ were not correlated with qualitative results, yielding unrealistically good scores for geometrically corrupted inputs. Our analysis shows that the model fails to segment KITTI-360 scenes, due to resolution differences: SemanticKITTI range images are twice as wide as those in KITTI-360. To address this, we replace RangeNet++ with RangeViT \cite{ando_rangevit_2023}, trained on SemanticKITTI range images whose spatial dimensions match those of KITTI-360. 

FSVD and FPVD are computed directly in the point cloud space using the LiDAR segmentation models MinkowskiNet \cite{choy2019minknet} and SPVCNN \cite{spvnas}, respectively. 
We argue that these metrics provide a more reliable measure of generation quality, as they operate directly on the point cloud and are not affected by projection discrepancies.

FLD is computed using the LIP-Loc model \cite{puligilla_lip-loc_2024}, which is trained in a CLIP-like manner for cross-modal localization between range images and camera images.

In addition to perceptual metrics, we compare the histograms of the bird’s-eye view point distribution using Jensen–Shannon Divergence (JSD). 
Finally, we compute the Minimum Matching Distance (MMD) \cite{ran_towards_2024}, which measures the Chamfer distance between each validation bird's-eye view voxelized scene and its closest generated counterpart.

\subsection{Comparison to state-of-the-art}
\newcommand\tabsetting{
\newcommand\rf[2]{\,{\scriptsize ##1\,'##2}}
}
\begin{table*}[htbp]

\caption{Quantitative Comparison of Unconditional Generation on KITTI-360.} 
\vspace{-0.2cm}
\label{tab:sota}
\centering
\tabsetting
\small
\resizebox{\linewidth}{!}{%

\begin{tabular}{lcccccccccccc}  
\toprule

&  &  \multicolumn{5}{c}{\textbf{Train-val}}  &  \multicolumn{5}{c}{\textbf{Val-only}} &  \\
\cmidrule(lr){3-7} \cmidrule(lr){8-12} 
 
 \multicolumn{1}{l}{\rotatebox{0}{\textbf{Method}}}   
  & \multicolumn{1}{c}{\textbf{Reference}}    
 & \multicolumn{1}{c}{\textbf{FRID}}    
 & \multicolumn{1}{c}{\textbf{FLD}}  
 & \multicolumn{1}{c}{\textbf{FSVD}}  
 & \multicolumn{1}{c}{\textbf{FPVD}}  
 & \multicolumn{1}{c}{\textbf{JSD}}    
   
 & \multicolumn{1}{c}{\textbf{FRID}}    
 & \multicolumn{1}{c}{\textbf{FLD}}  
 & \multicolumn{1}{c}{\textbf{FSVD}}  
 & \multicolumn{1}{c}{\textbf{FPVD}}  
 & \multicolumn{1}{c}{\textbf{JSD}}    
 & \multicolumn{1}{c}{\textbf{MMD}}  
 \\
   & & {\footnotesize $\times 10^{0}$} & {\footnotesize $\times 10^{-1}$}  & {\footnotesize $\times 10^{0}$} & 
{\footnotesize $\times 10^{0}$} & {\footnotesize $\times 10^{-2}$}  & {\footnotesize $\times 10^{0}$} & {\footnotesize $\times 10^{-1}$} & {\footnotesize $\times 10^{0}$} & {\footnotesize $\times 10^{0}$} & {\footnotesize $\times 10^{-2}$} & {\footnotesize $\times 10^{-5}$}\\
 \midrule

UltraLiDAR & \cite{xiong_ultralidar_2023}\rf{cvpr}{23}  & - & - & 71.69 & 64.78 & 74.72  & - & - & 73.59 & 65.83 & 74.72 &  123.30 \\

LiDM & \cite{ran_towards_2024}\rf{cvpr}{24}  & 42.31 & 9.52 & 16.54 & 17.60 & 18.87  & 47.33 & 10.19 & 16.01 & 17.36 & 19.17 &  11.32 \\

LiDM w/ APE& \cite{ran_towards_2024}\rf{cvpr}{24}  & 37.01 & 6.12 & 13.79 & 13.82 & 11.24  & 42.09 & 9.76 & 13.68 & 13.86 & 11.69 &  9.95 \\

R2DM  & \cite{nakashima_lidar_2024}\rf{icra}{24}      & 9.84 & 4.04 & \underline{10.99} & \underline{11.99} & \underline{4.97}  &  15.54 & \underline{7.89} & \underline{12.67} & \underline{13.21} & \underline{5.78}  & \underline{8.50} \\

R2Flow   & \cite{nakashima_fast_2025}\rf{icra}{25}   & \textbf{3.89} & \underline{3.82} & 18.48 & 18.53 & 5.17  & \underline{8.87} & 8.36 & 20.80 & 20.27 & 5.97 &  \textbf{7.84}  \\
\midrule
\textbf{\ours} & ours & \underline{4.11} & \textbf{2.06} & \textbf{7.76} & \textbf{9.39} & \textbf{4.84} & \textbf{8.46} & \textbf{6.34} & \textbf{9.83} & \textbf{11.00} & \textbf{5.67}  & 8.72 \\

\bottomrule
\end{tabular}
}

\begin{tablenotes}[para]
    \footnotesize Best results are shown in \textbf{bold}, and second-best results are \underline{underlined}. FRID and FLD measure generation quality in the range image level, FSVD and FPVD in the point-cloud space, and JSD and MMD in the bird’s-eye view. APE stands for Absolute Positional Encoding.
\end{tablenotes}

\end{table*}

\begin{figure*}[t!]
    \centering
    \begin{tabular}{cccc}
        \begin{minipage}{0.22\textwidth}
            \centering
            \includegraphics[width=\linewidth]{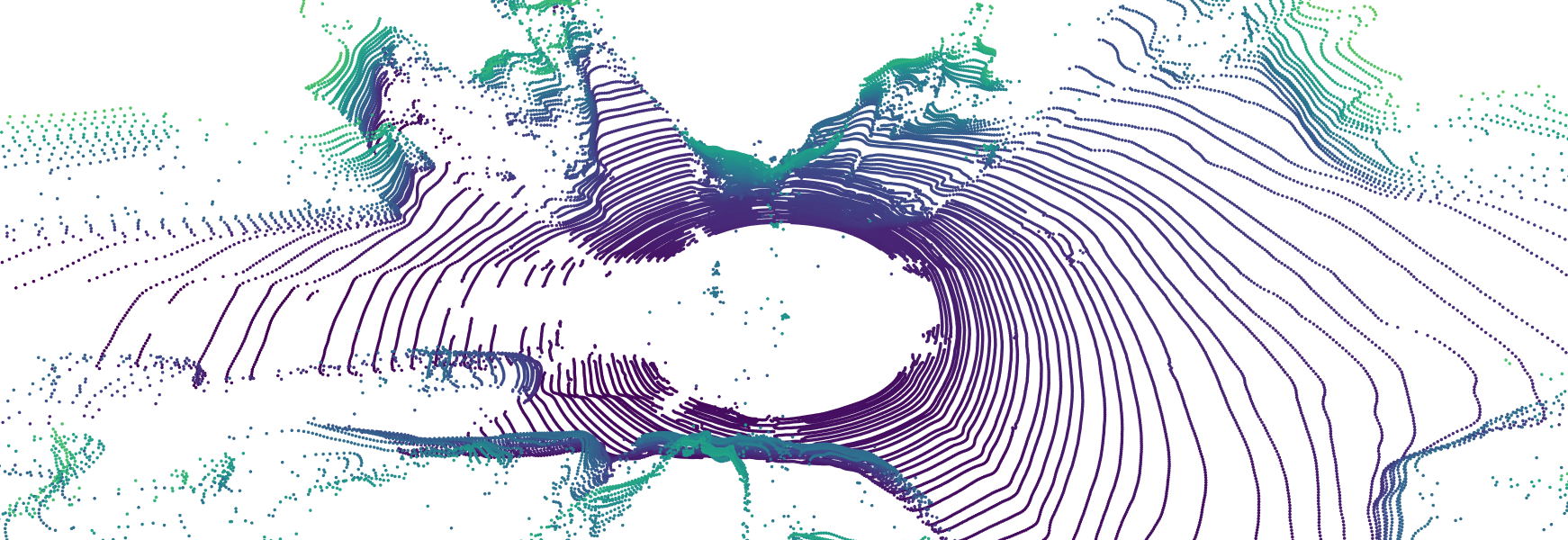} \\[0.5ex]
            \includegraphics[width=\linewidth]{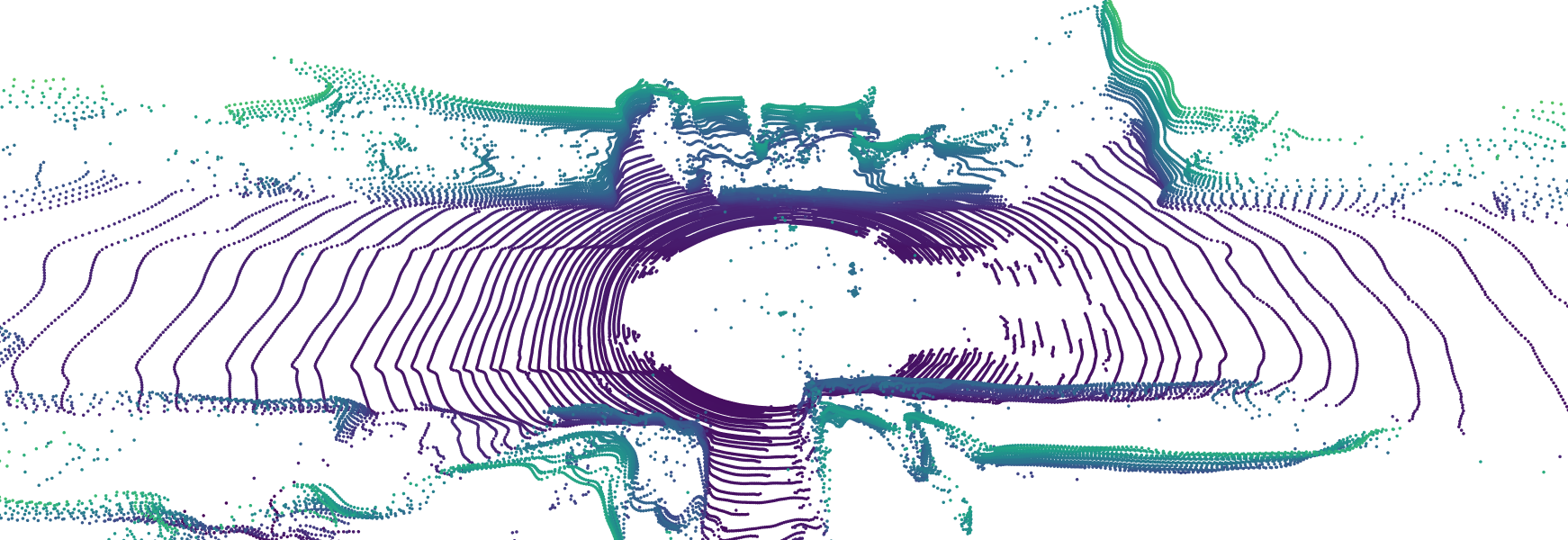} \\[0.5ex]
            \small{LiDM~\cite{ran_towards_2024}}
        \end{minipage} &
        \begin{minipage}{0.22\textwidth}
            \centering
            \includegraphics[width=\linewidth]{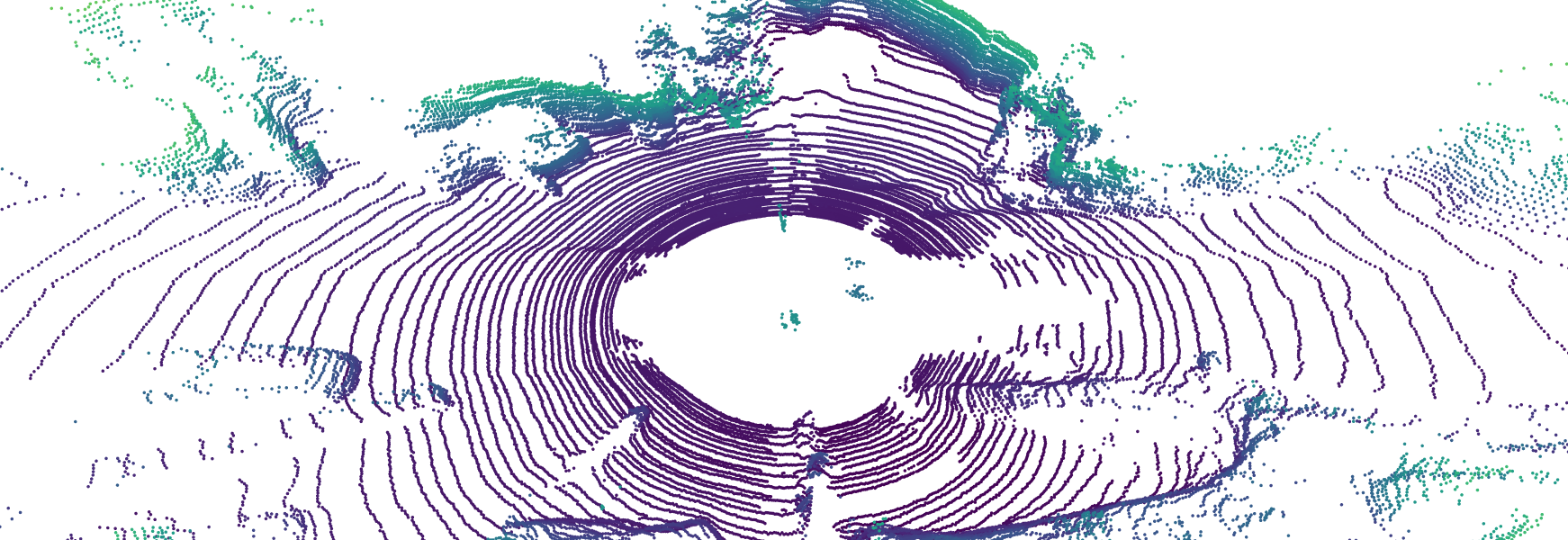} \\[0.5ex]
            \includegraphics[width=\linewidth]{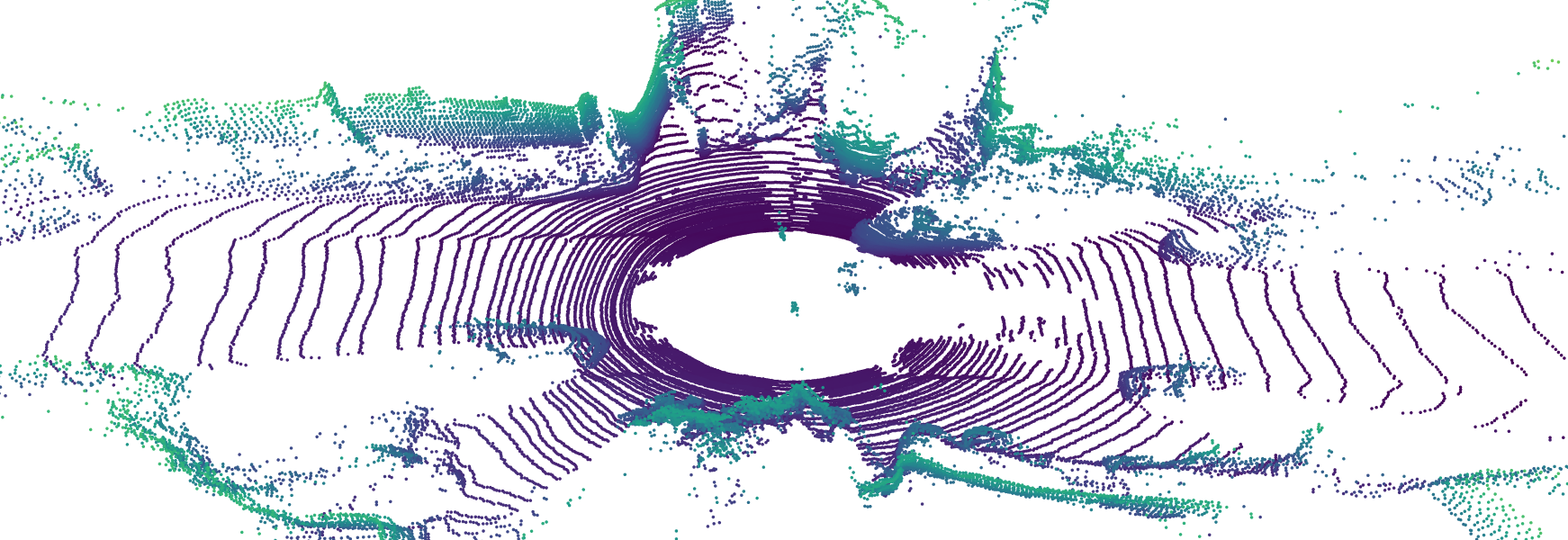} \\[0.5ex]
            \small{R2DM~\cite{nakashima_lidar_2024}}
        \end{minipage} &
        \begin{minipage}{0.22\textwidth}
            \centering
            \includegraphics[width=\linewidth]{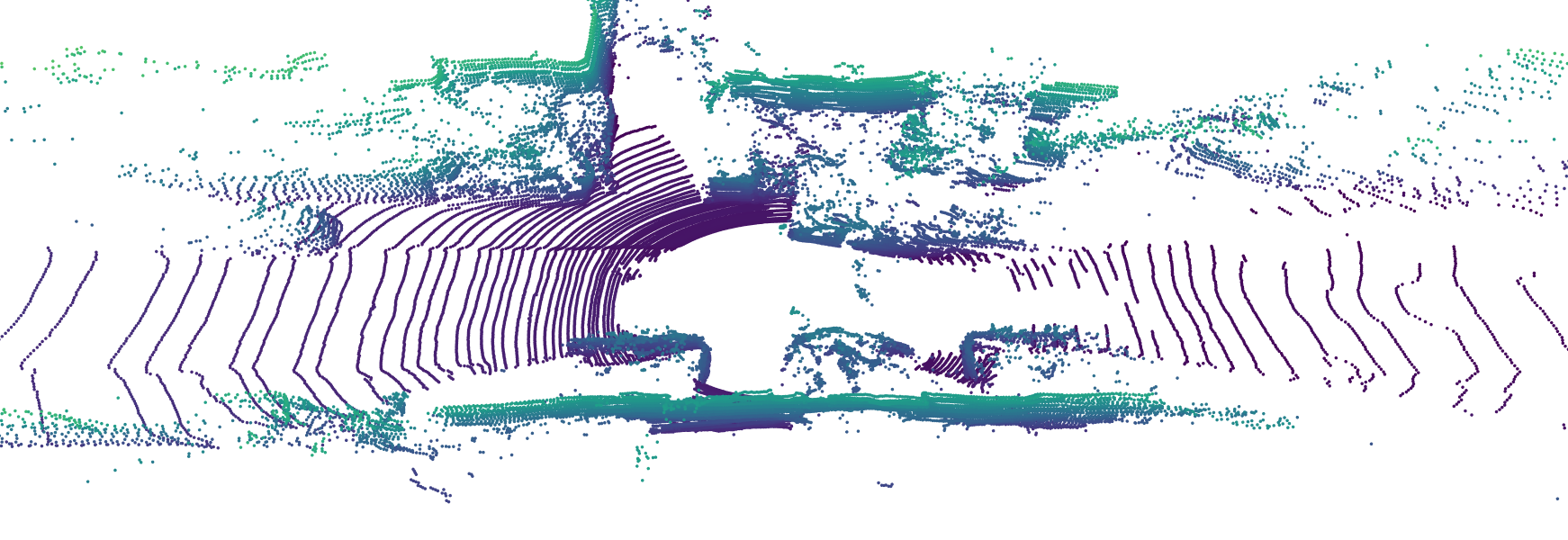} \\[0.5ex]
            \includegraphics[width=\linewidth]{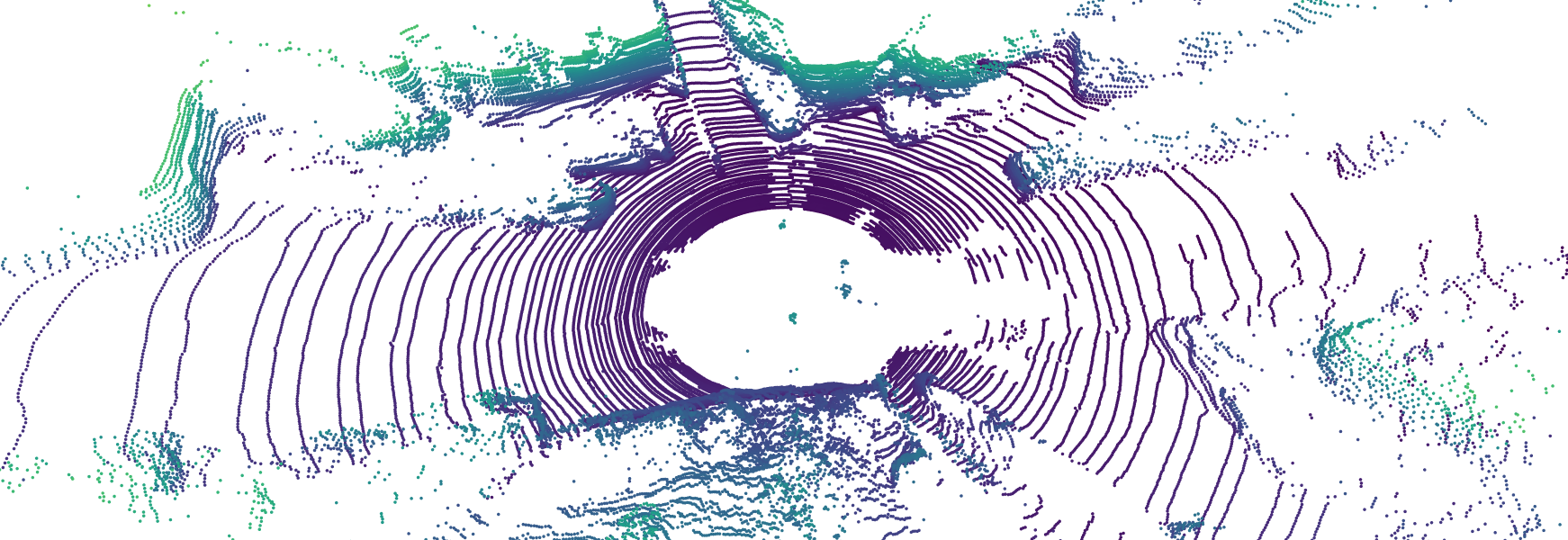} \\[0.5ex]
            \small{R2Flow~\cite{nakashima_fast_2025}}
        \end{minipage} &
        \begin{minipage}{0.22\textwidth}
            \centering
            \includegraphics[width=\linewidth]{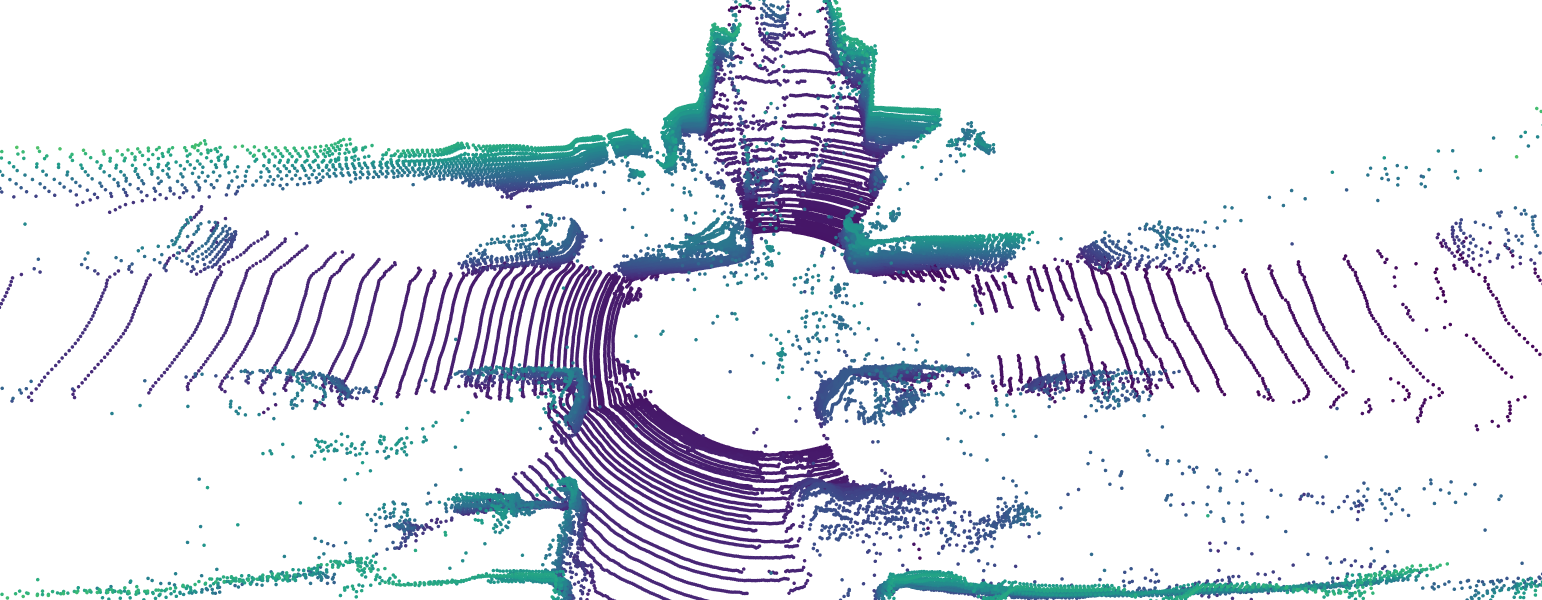} \\[0.5ex]
            \includegraphics[width=\linewidth]{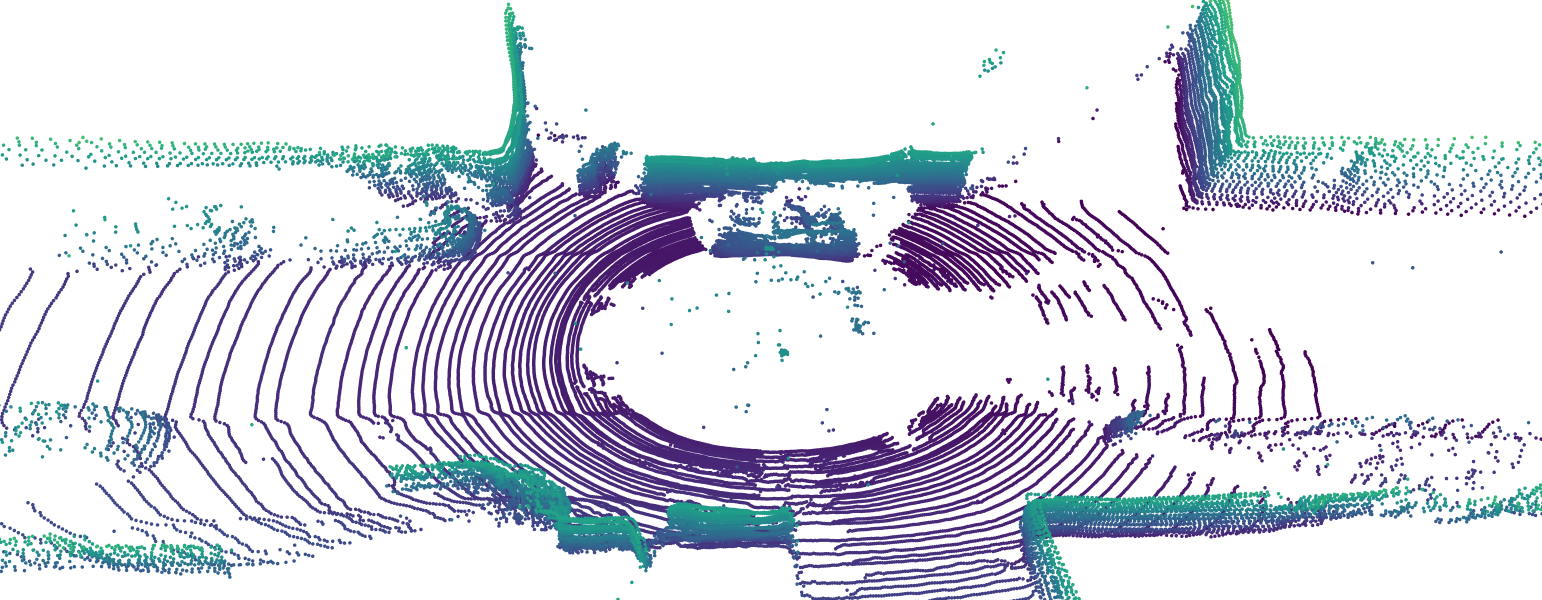} \\[0.5ex]
            \small{\textbf{\ours}}
        \end{minipage} \\
    \end{tabular}
    \caption{\textbf{Qualitative comparison of unconditional generations.} Our model generates high-quality point clouds with diverse and realistic objects.}
    \label{fig:qual}
\end{figure*}


\begin{figure*}[t]
    \centering
    \begin{minipage}{0.24\textwidth}
        \begin{tikzpicture}
            \node[anchor=south west, inner sep=0] (img1) at (0,0) 
                {\includegraphics[width=\linewidth]{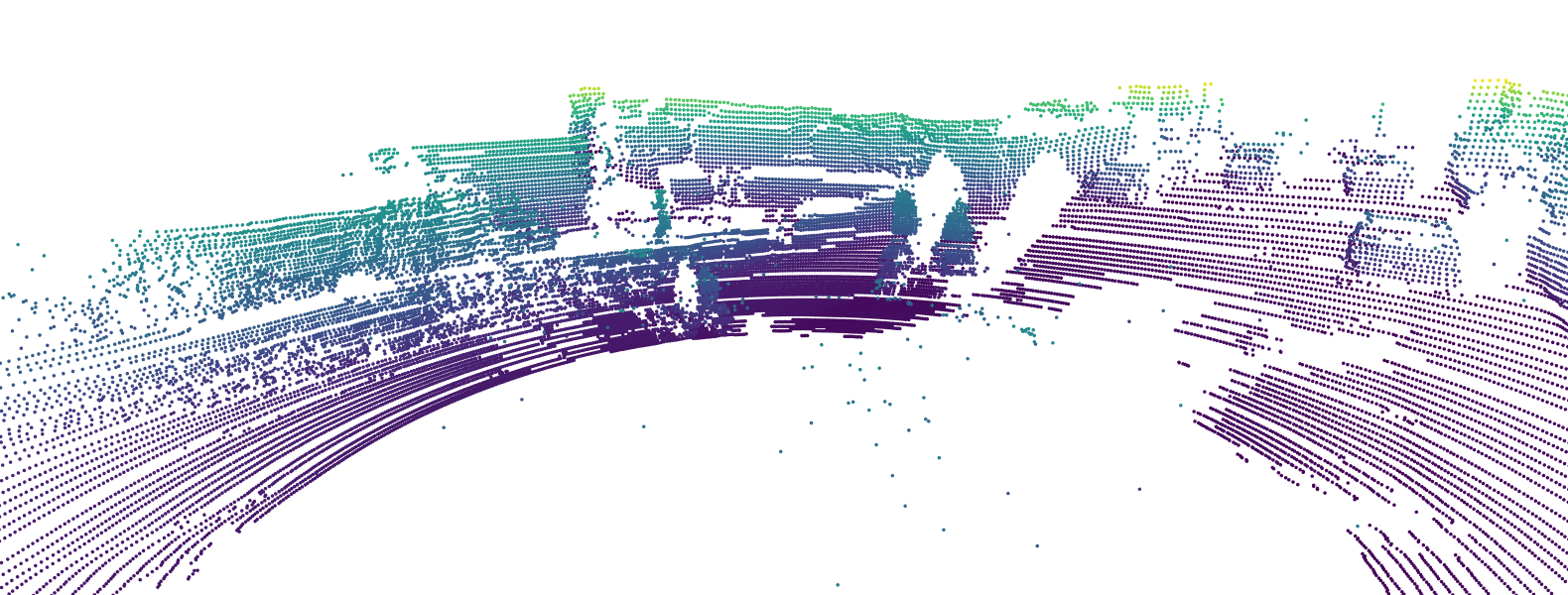}};
            \begin{scope}[x={(img1.south east)},y={(img1.north west)}]
                \draw[red, thick] (0.45, 0.49) circle (4pt); 
                \draw[red, thick] (0.59, 0.59) circle (5pt); 
            \end{scope}
        \end{tikzpicture}
    \end{minipage}
    \hfill
    \begin{minipage}{0.24\textwidth}
        \begin{tikzpicture}
            \node[anchor=south west, inner sep=0] (img2) at (0,0) 
                {\includegraphics[width=\linewidth]{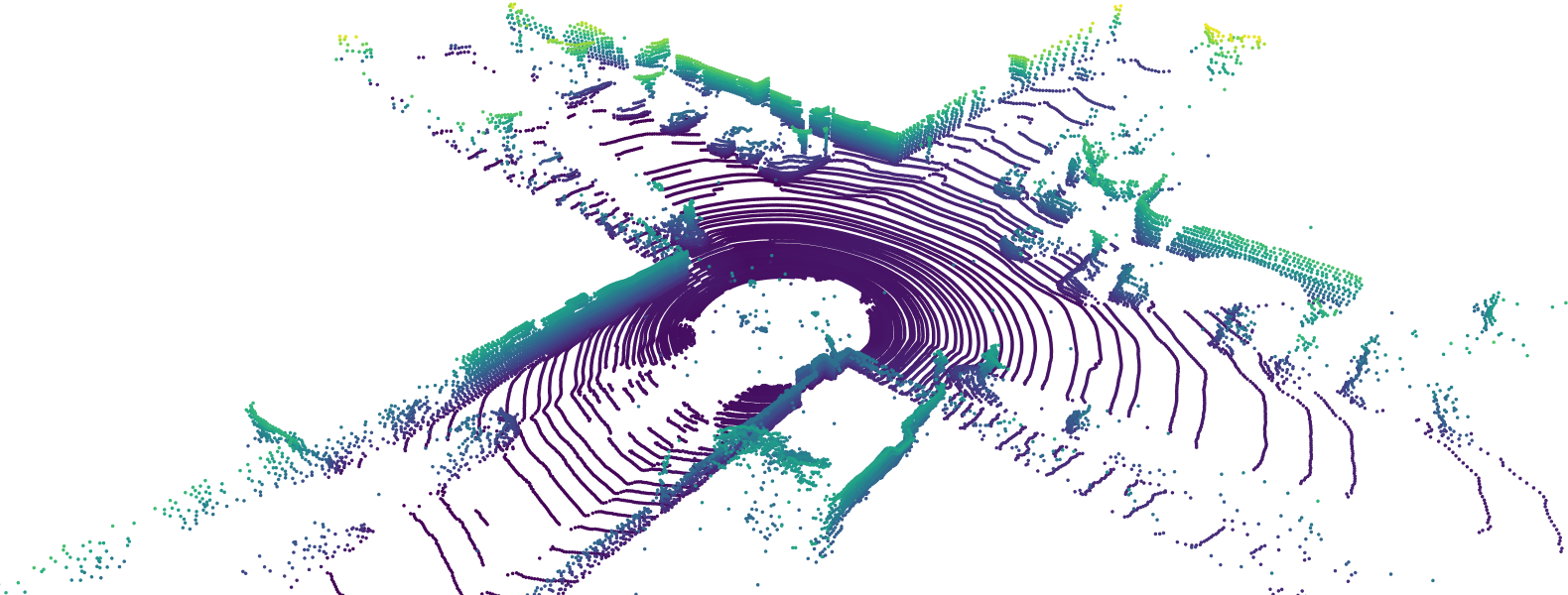}};
            \begin{scope}[x={(img2.south east)},y={(img2.north west)}]
            \end{scope}
        \end{tikzpicture}
    \end{minipage}
    \hfill
    \begin{minipage}{0.24\textwidth}
        \begin{tikzpicture}
            \node[anchor=south west, inner sep=0] (img3) at (0,0) 
                {\includegraphics[width=\linewidth]{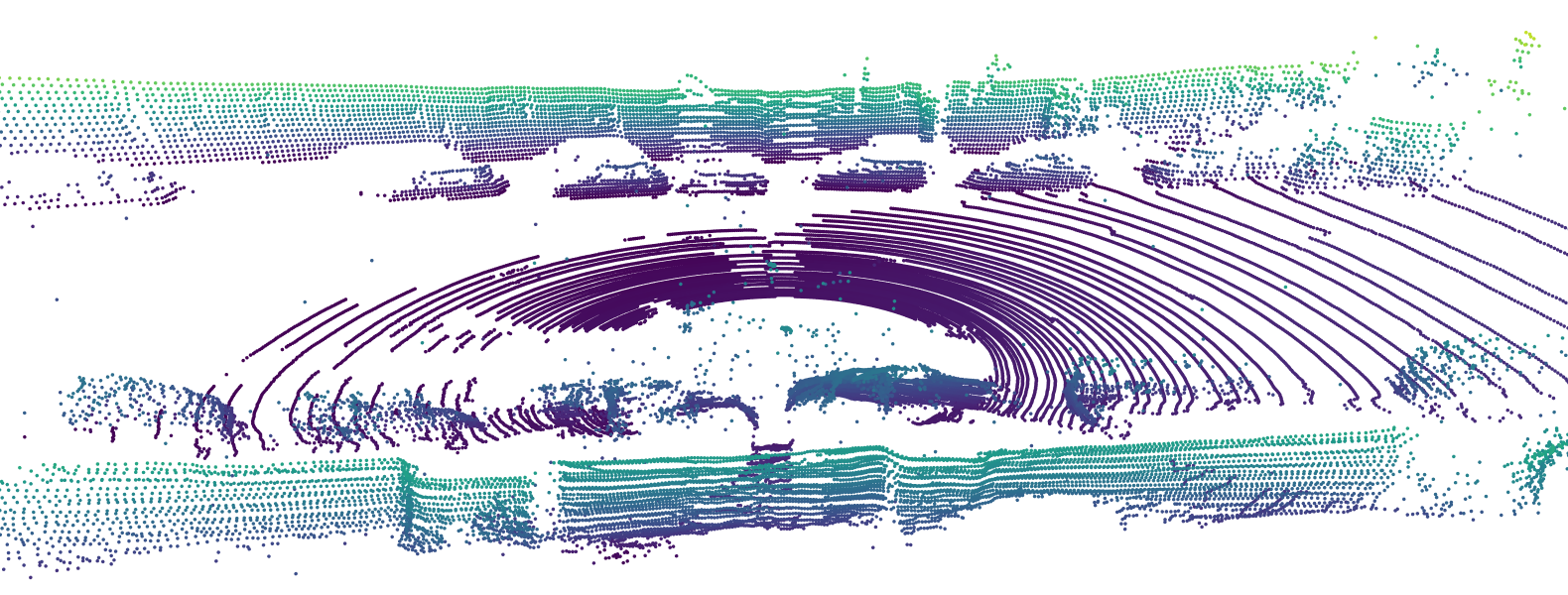}};
            \begin{scope}[x={(img3.south east)},y={(img3.north west)}]
            \end{scope}
        \end{tikzpicture}
    \end{minipage}
    \hfill
    \begin{minipage}{0.24\textwidth}
        \begin{tikzpicture}
            \node[anchor=south west, inner sep=0] (img4) at (0,0) 
                {\includegraphics[width=\linewidth]{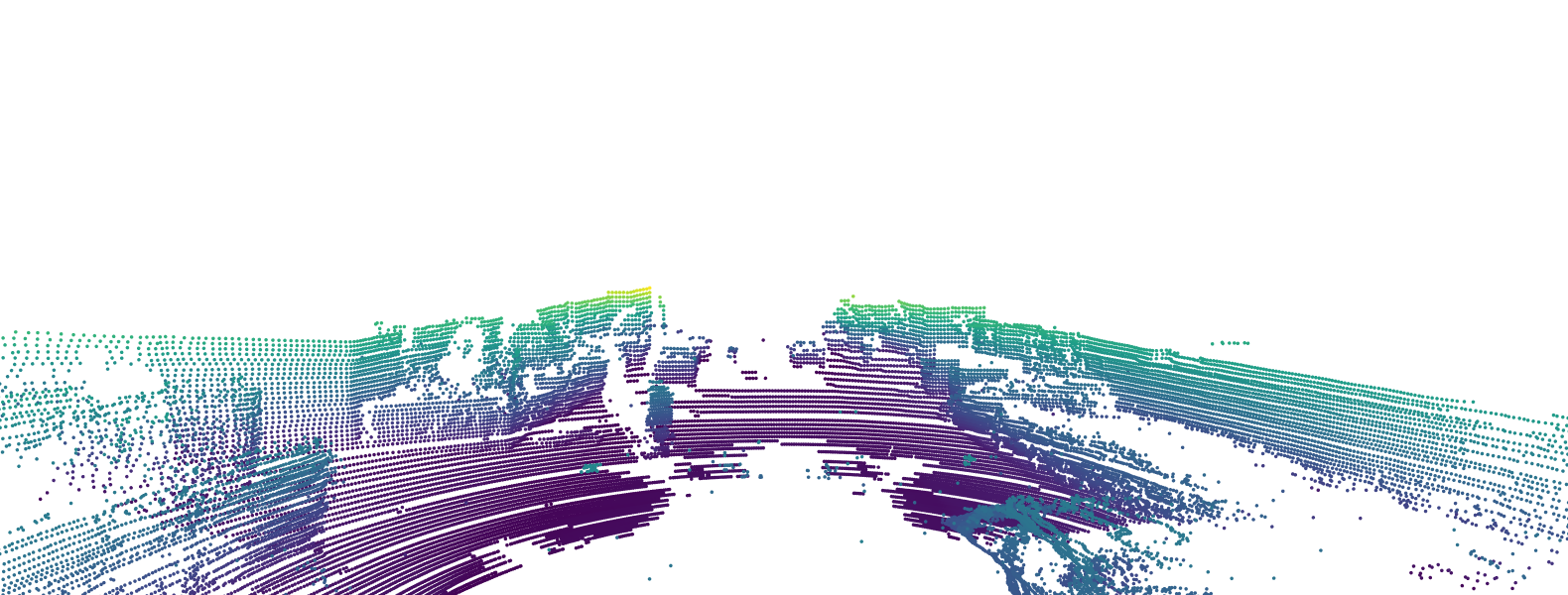}};
            \begin{scope}[x={(img4.south east)},y={(img4.north west)}]
                \draw[red, thick] (0.42, 0.3) circle (4pt);
            \end{scope}
        \end{tikzpicture}
    \end{minipage}
    \caption{\textbf{Qualitative samples from \ours.}
    We generate diverse scenes with cars, pedestrians and cyclist while faithfully capturing the LiDAR pattern. }
    \label{fig:ours_qual}
\end{figure*}
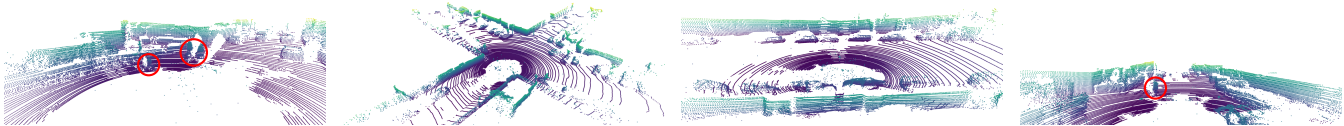

We compare our method against SOTA scene generation methods. We generate samples using a comparable number of steps: 250 for LiDM~\cite{ran_towards_2024} and 256 for R2DM~\cite{nakashima_lidar_2024} and R2Flow~\cite{nakashima_fast_2025}, following their official repositories. For fairness, we also use 256 steps for our model. For UltraLiDAR \cite{xiong_ultralidar_2023}, we use the publicly available samples released online. We generate our samples following \cite{ma_sit_2024} using the stochastic differential equation with the Euler-Maruyama solver.

To ensure a fair comparison, we follow~\cite{nakashima_fast_2025} and train LiDM \cite{ran_towards_2024} with a learnable absolute positional encoding inserted after the first UNet convolution.
This stabilizes orientation in generated scenes, allowing evaluation with non-rotation-invariant metrics (i.e. JSD, MMD) and improving performance on rotation-invariant metrics.

For each method, we generate 10,000 samples and evaluate them against the full dataset of real scenes (train-val evaluation), as is common in generative modeling \cite{nakashima_fast_2025}. 
For completeness, we also report results evaluated against the validation set (val-only evaluation).
For MMD, which requires computing the expensive Chamfer distance between all pairs of point clouds, we use 2,000 generated samples and evaluate them against 1,000 validation scenes, as in \cite{ran_towards_2024}.
The VAE alignment and end-to-end training require 48h and 70h, respectively, using 8 NVIDIA A100 GPUs.  \ours achieves an average inference speed of 0.6 s/scene, generating 10k samples in 1.45h on a single A100 with a batch size of 64.

We present quantitative results in Table~\ref{tab:sota} and qualitative comparisons in Figure~\ref{fig:qual}. For range-image metrics, \ours performs on par with the state of the art on FRID, while achieving a significant improvement on the FLD metric. For point-based metrics, which are particularly important given the ultimate goal of generating high-quality, realistic point clouds, our method surpasses previous approaches.
On bird’s-eye-view metrics, \ours improves the JSD score while achieving performance comparable to the state of the art on MMD.
Importantly, we achieve strong performance across both point and range metrics while other works show higher variance. We suspect that \cite{xiong_ultralidar_2023}'s low score likely results from its voxel-based architecture.
We present further qualitative results from our model in Figure \ref{fig:ours_qual}.

\begin{table}[htbp]
\caption{Effect of RGB Image Pretrained Priors and VAE Alignment.}
\vspace{-0.2cm}
\centering
\small
\resizebox{\linewidth}{!}{%
\begin{tabular}{ccccccccc}
\toprule

 \multicolumn{1}{c}{\rotatebox{0}{\textbf{RGB}}}   
 & \multicolumn{1}{c}{\textbf{VAE}}  
 & \multicolumn{1}{c}{\textbf{FRID}}    
 & \multicolumn{1}{c}{\textbf{FLD}}  
 & \multicolumn{1}{c}{\textbf{FSVD}}  
 & \multicolumn{1}{c}{\textbf{FPVD}}  
 & \multicolumn{1}{c}{\textbf{JSD}}    

 & \multicolumn{1}{c}{\textbf{MMD}}  
 \\
\textbf{Priors}  & \textbf{Align.} & {\footnotesize $\times 10^{0}$} & {\footnotesize $\times 10^{-1}$} & 
{\footnotesize $\times 10^{0}$} & {\footnotesize $\times 10^{0}$} & {\footnotesize$\times 10^{-2}$} &  {\footnotesize $\times 10^{-5}$}\\
 \midrule
\xmark & \xmark  & 13.44 & 4.98 & 9.97 & 11.21 & 5.08 &   8.99 \\
\cmark & \xmark &   12.81 & 4.52 & 10.13 & 11.39 & 4.98 &   8.87  \\
\cmark & \cmark &  \textbf{7.12} & \textbf{3.26} & \textbf{8.55} & \textbf{10.37 }& \textbf{4.70 }&   \textbf{8.76} \\
\bottomrule
\end{tabular}
}
\label{tab:abl_vae} 

\begin{tablenotes}[para]
    \footnotesize Direct use of RGB priors does not improve performance, whereas VAE alignment yields improvements across all metrics. 
\end{tablenotes}
\end{table}

\subsection{Ablations}

For faster experimentation in ablation studies, we first train the VAE from scratch for 400k steps during the alignment stage, and then perform end-to-end training of both the VAE and SiT for another 400k steps.
All ablation results are reported under the train-val evaluation setting.

\textbf{Effect of RGB Priors and VAE Alignment.} 
In this series of ablation experiments, we evaluate the impact of (i) leveraging priors from a backbone pretrained on a large-scale image dataset and (ii) incorporating our proposed VAE alignment step to better transfer these priors. To this end, we consider three settings: 

\begin{enumerate} 
    \item A base model where the VAE is trained from scratch in the first stage; then by keeping the VAE frozen the SiT is trained from scratch with random initialization in the second stage. 
    \item  The same as (1), except in the second stage SiT is initialized with RGB image pretrained weights instead of random initialization.
    \item VAE alignment is first applied by training the VAE from scratch to adapt it to pretrained RGB image priors; in the second stage, the VAE is kept frozen and SiT is trained, initialized with RGB image pretrained weights.
\end{enumerate} 

We note that these experiments do not employ end-to-end training. Instead, we follow the classical two-stage paradigm as in~\cite{ran_towards_2024,hu_rangeldm_2024}, where the VAE is first trained independently and then kept frozen during the second stage while training the FM model.

As shown in Table~\ref{tab:abl_vae}, using RGB priors alone does not improve performance, whereas the VAE alignment step leads to consistent improvements across all metrics. We conclude that the VAE alignment step reduces the gap with respect to the ImageNet-trained SiT input space, enabling more effective use of RGB prior knowledge. 

\begin{figure*}[ht]
\centering
\begin{tabular}{lcc}
  \centering
\raisebox{5mm}{\multirow{1}{*}{\rotatebox[origin=c]{90}{\footnotesize GT}}} & 
\includegraphics[width=0.46\linewidth]{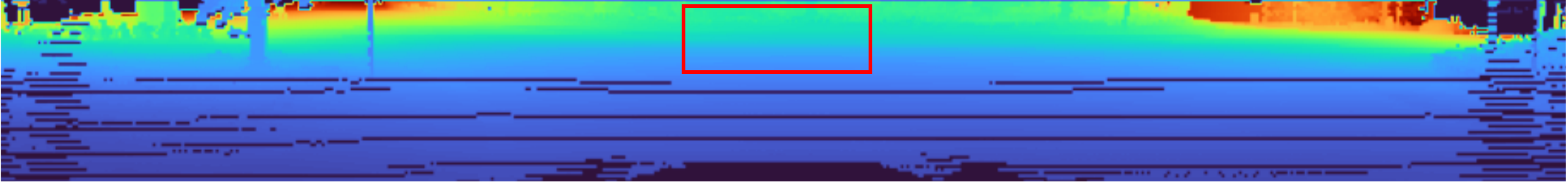} & 
\includegraphics[width=0.46\linewidth]{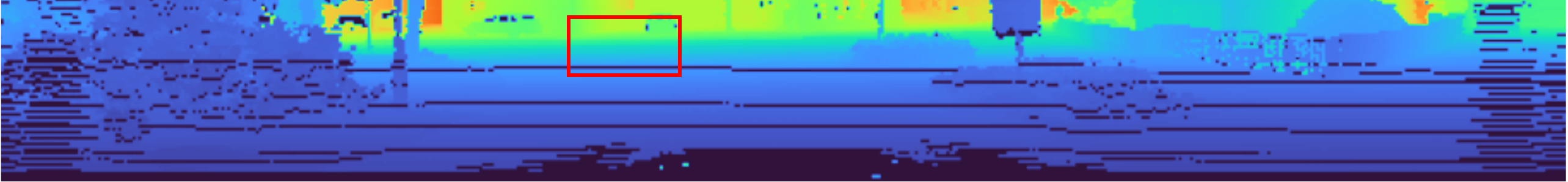}\\ 
  
\raisebox{5mm}{\multirow{1}{*}{\rotatebox[origin=c]{90}{ \footnotesize Gen}}} & 
\includegraphics[width=0.46\linewidth]{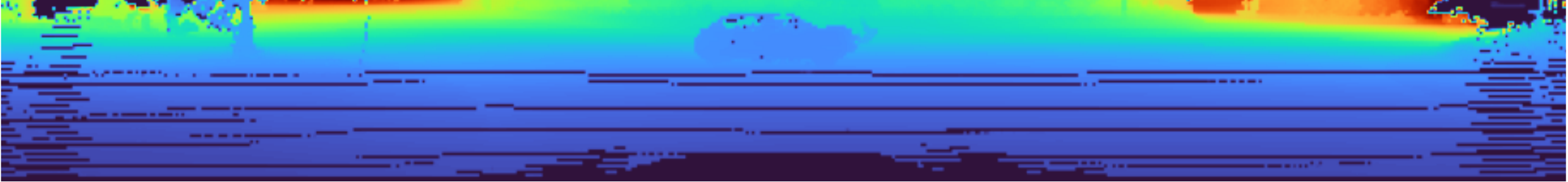} & 
\includegraphics[width=0.46\linewidth]{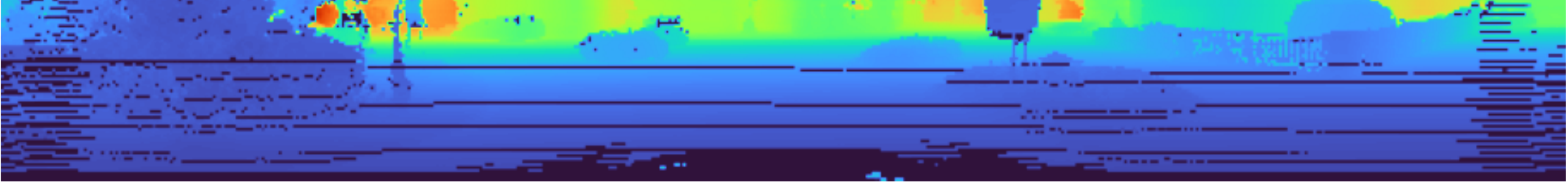} \\
    
& \cropcenter{0.22\linewidth}{\detokenize{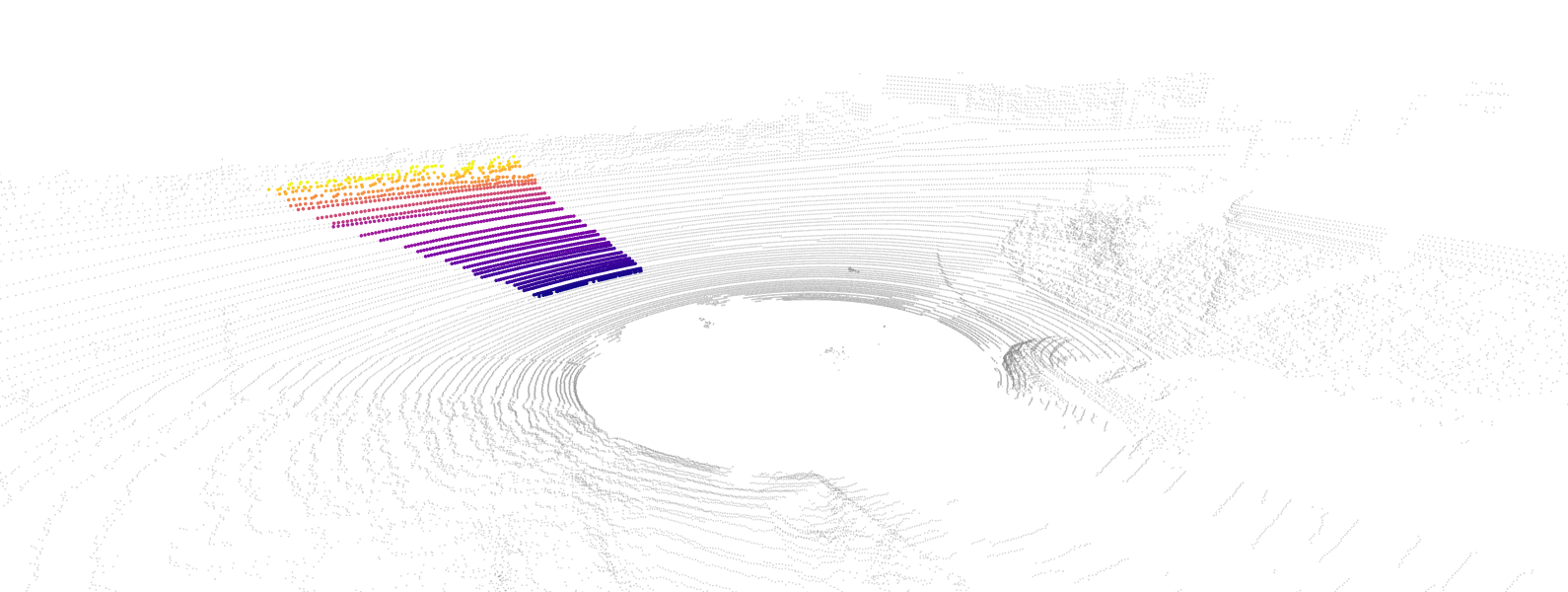}}  \cropcenter{0.22\linewidth}{\detokenize{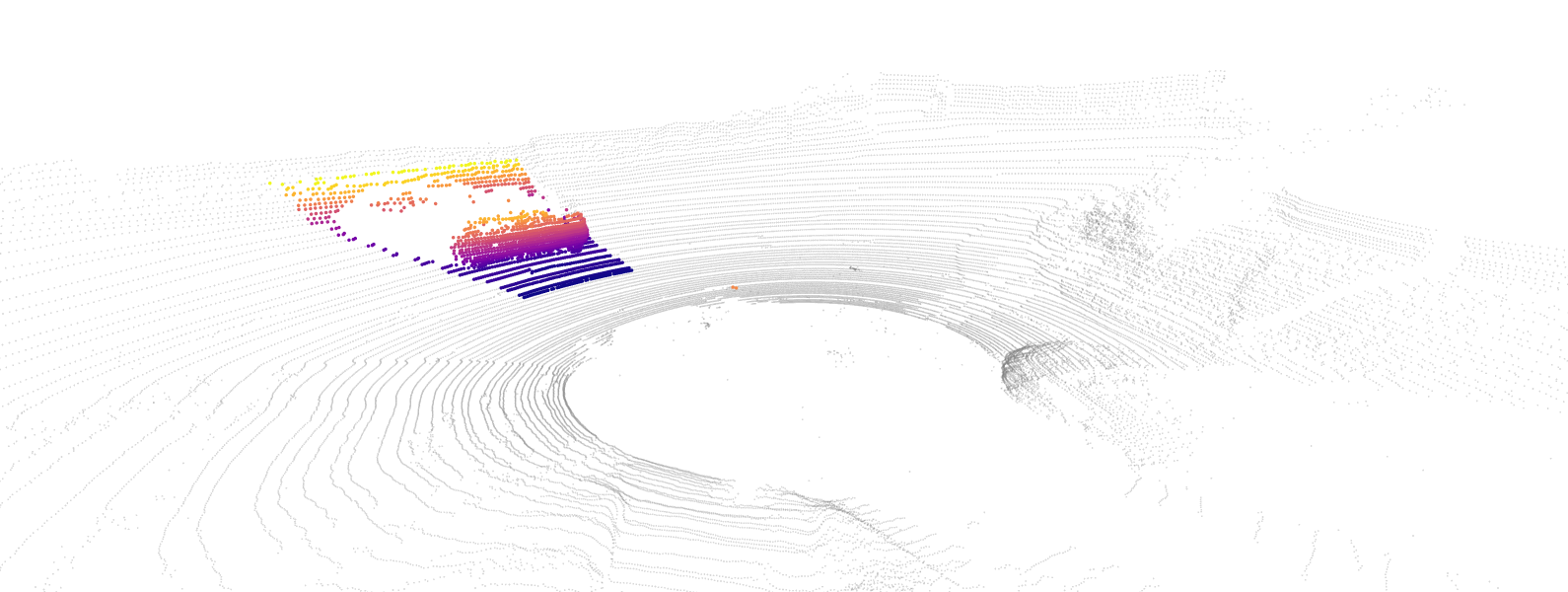}} & 
\cropcenter{0.22\linewidth}{\detokenize{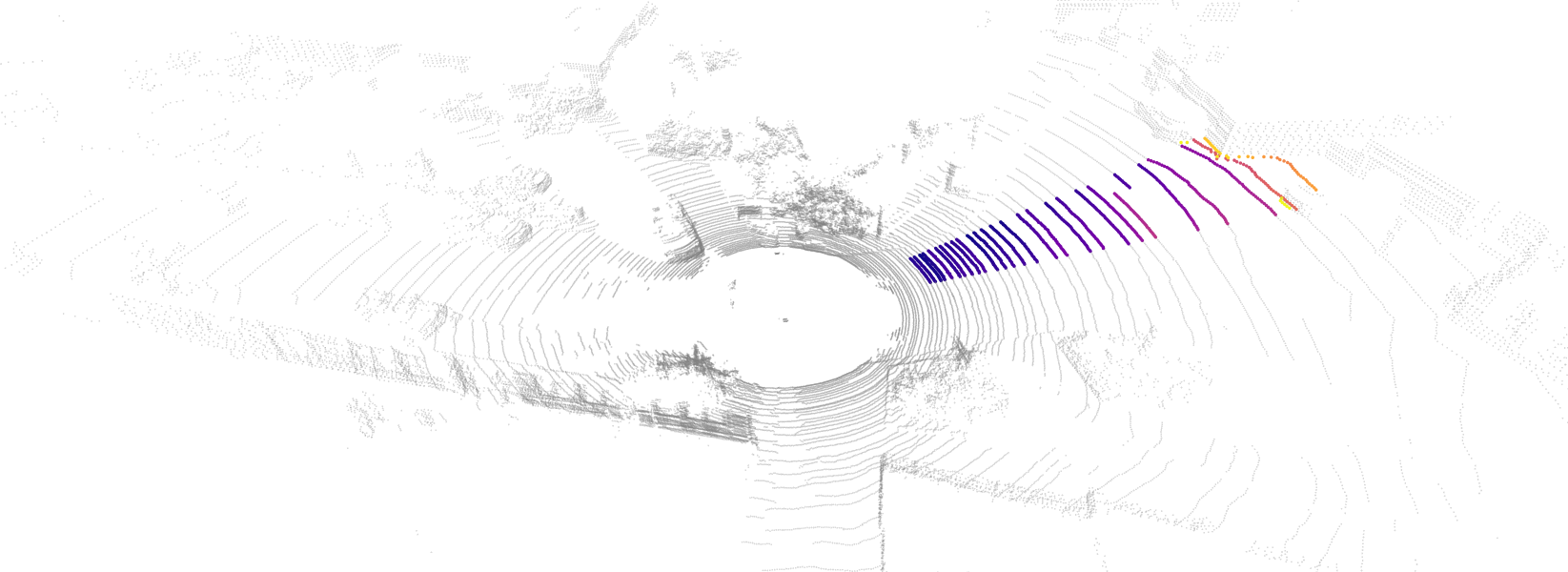}}
\cropcenter{0.22\linewidth}{\detokenize{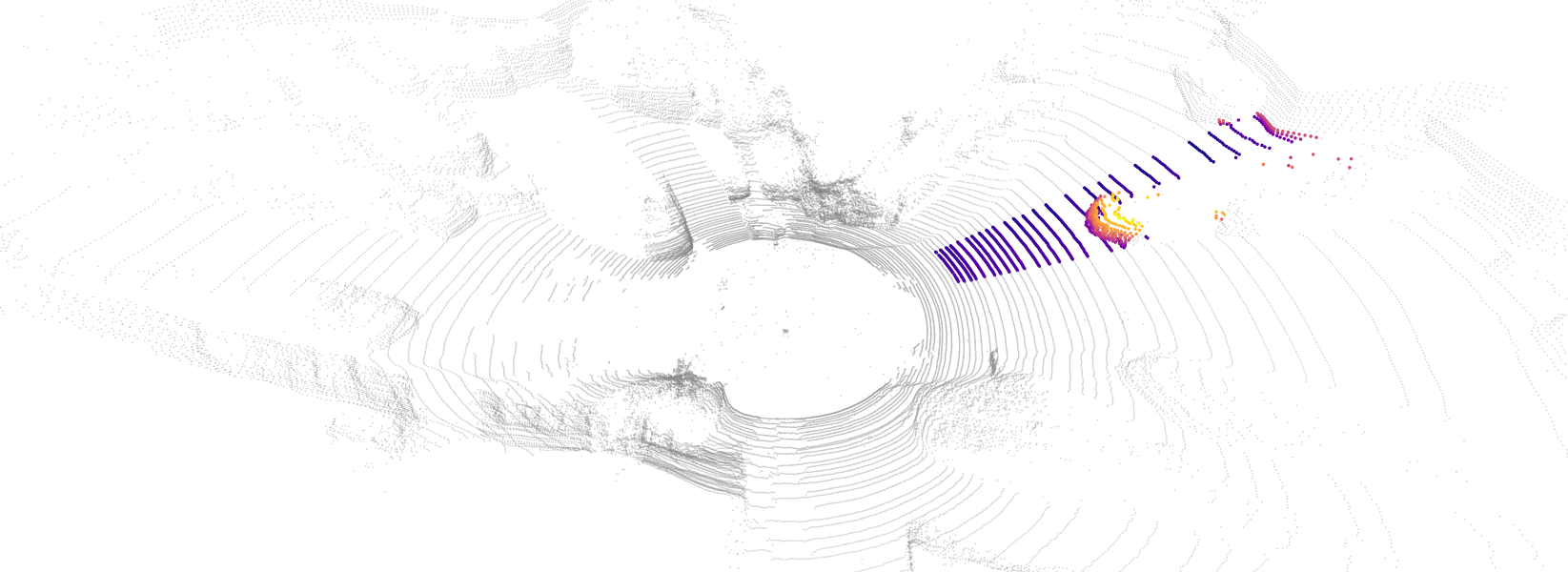}} \\
& \footnotesize GT PC {~~~~~~~~~~~~~~~~~~~~~~~~~~~~~} \footnotesize Gen PC & \footnotesize GT PC  {~~~~~~~~~~~~~~~~~~~~~~~~~~~~~}  \footnotesize Gen PC   \\

\end{tabular}
\caption{\textbf{Test-time conditioning with 3D representations for object inpainting.} A user-specified mask (red) is drawn on top of the ground truth range image (center-cropped for visibility). The real scene is transported to the Gaussian distribution, then the user defined area is filled with noise. During the denoising process only the masked area is supervised with features corresponding to the target object class. The scenes are also presented projected to 3D points, colored according to the intersection with the frustum of the projected mask.}
\label{fig:inpaint}
\vspace{-0.2cm}
\end{figure*}
\begin{table}[htbp]
\caption{3D Representation Alignment and End-to-end Training.}
\vspace{-0.2cm}
\centering
\small
\resizebox{\linewidth}{!}{%
\begin{tabular}{ccccccc}
\toprule
 \multicolumn{1}{c}{\textbf{2nd Stage}}  
 & \multicolumn{1}{c}{\textbf{FRID}}    
 & \multicolumn{1}{c}{\textbf{FLD}}  
 & \multicolumn{1}{c}{\textbf{FSVD}}  
 & \multicolumn{1}{c}{\textbf{FPVD}}  
 & \multicolumn{1}{c}{\textbf{JSD}}   

 & \multicolumn{1}{c}{\textbf{MMD}}  
  \\
 \textbf{Alignment} & {\footnotesize $\times 10^{0}$} & {\footnotesize $\times 10^{-1}$} & 
{\footnotesize $\times 10^{0}$} & {\footnotesize $\times 10^{0}$} & {\footnotesize$\times 10^{-2}$} &  {\footnotesize $\times 10^{-5}$}\\
 \midrule
 No       & 7.12 & 3.26 & 8.55 & 10.37 & 4.70 &   8.76 \\
 Vanilla  & 7.23 & 3.28 & 8.52 & 10.25 & 4.64 & 8.81 \\
 Joint    & \textbf{4.14} & \textbf{2.23} & \textbf{7.96} & \textbf{9.69} & \textbf{4.55} &  \textbf{8.41} \\
\bottomrule
\end{tabular}
}
\hspace{-0.3cm}
\begin{tablenotes}[para]
    \footnotesize 
     We first perform the VAE alignment step and use its weights in the subsequent experiments. \textit{No alignment:} the VAE is kept frozen and SiT is trained with RGB image pretrained weights. \textit{Vanilla alignment:} the VAE is kept frozen and 3D representation guidance is propagated to the flow model. \textit{Joint alignment:} end-to-end training with 3D guidance propagated to both the flow model and the VAE encoder.
\end{tablenotes}
\label{tab:abl_repas}
\end{table}

\textbf{Effect of 3D Alignment and End-to-end Training.}
In this series of experiments, we study the impact of 3D representation alignment and end-to-end training. 
In the previous section, we leveraged RGB priors from pretrained models by first performing VAE alignment and then, in a second stage, freezing the VAE while training the FM model SiT initialized with RGB image pretrained weights. Since this setting does not incorporate any guidance from 3D representations during second stage, we refer to it as \textit{no alignment}. 
The next experiment, \textit{vanilla alignment}, also keeps the VAE frozen but propagates 3D representation guidance only to the FM model. 
Finally, \textit{joint alignment} corresponds to our end-to-end training setup, where 3D representation guidance is propagated to both the FM model and the VAE encoder.

Table~\ref{tab:abl_repas} shows that end-to-end training with 3D representations substantially improves generation quality across all metrics.

\begin{table}[t]
\caption{Comparison of UNet and SiT.}
\vspace{-0.2cm}
\label{tab:abl_arch}
\centering
\small
\resizebox{\linewidth}{!}{%
\begin{tabular}{ccccccc}
\toprule
  \multirow{2}{*}{\textbf{Backbone}}   
 & \multicolumn{1}{c}{\textbf{FRID}}    
 & \multicolumn{1}{c}{\textbf{FLD}}  
 & \multicolumn{1}{c}{\textbf{FSVD}}  
 & \multicolumn{1}{c}{\textbf{FPVD}}  
 & \multicolumn{1}{c}{\textbf{JSD}}    
 & \multicolumn{1}{c}{\textbf{MMD}}  
 \\
  &  {\footnotesize $\times 10^{0}$} & {\footnotesize $\times 10^{-1}$} & 
{\footnotesize $\times 10^{0}$} & {\footnotesize $\times 10^{0}$} & {\footnotesize$\times 10^{-2}$}  & {\footnotesize $\times 10^{-5}$}\\
 \midrule
UNet of \cite{ran_towards_2024}   &  7.70 & \textbf{3.09} & \textbf{7.80} & 9.61 & 4.64   & \textbf{8.83}  \\ 
SiT-B/2 &  \textbf{6.08} & 3.17 & 7.89 & \textbf{8.86} & \textbf{4.52} & 8.95 \\

\bottomrule

\end{tabular}
}

\begin{tablenotes}[para]
    \footnotesize
    \hspace{-0.2cm} SiT achieves comparable or better results with twice fewer parameters.
\end{tablenotes}
\vspace{-0.5cm}

\end{table}

\textbf{Comparison of UNet and SiT.} UNet is one of the most widely used architectures for implementing diffusion and FM models~\cite{ran_towards_2024,nakashima_lidar_2024, zyrianov2022learning,hu_rangeldm_2024}. 
To provide a comparison with SiT, we use the UNet architecture from LiDM~\cite{ran_towards_2024}. Following~\cite{tian_u-repa_2025}, we supervise the representations of its middle block with projected ScaLR \cite{puy2024three} features for an end-to-end training \NS{without VAE Alignment step}.  
We note that UNet has roughly twice as many parameters as our SiT backbone (266M vs. 130M). 
Nevertheless, Table~\ref{tab:abl_arch} shows that SiT-B/2 achieves comparable or better results with fewer parameters.

\subsection{Test-Time Conditioning with 3D Representations}

We explore inference time guidance using the alignment loss without retraining.
Given a feature grid of a scene (such as Figure~\ref{fig:scalr_features}), generation can be guided toward consistency with these features by backpropagating the gradient of the alignment loss with respect to the latent vector $z_t$. 
The loss is computed between the SiT's internal representation and the conditional features, and combined with the SiT velocity vector to transport the latent toward a LiDAR scene matching the target features.  

\begin{figure}[t!]
    \centering
    \begin{tabular}{ccc}
        \includegraphics[width=0.29\columnwidth]{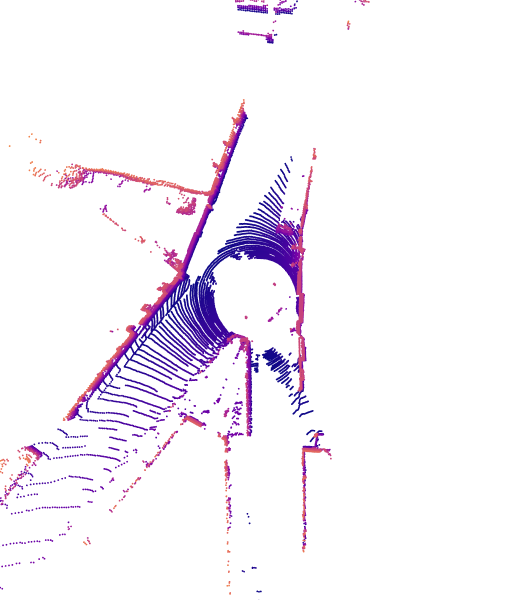} &
        \includegraphics[width=0.29\columnwidth]{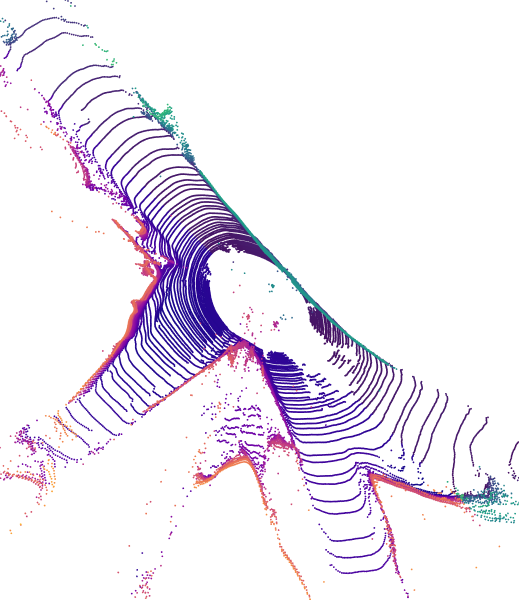} &
        \includegraphics[width=0.29\columnwidth]{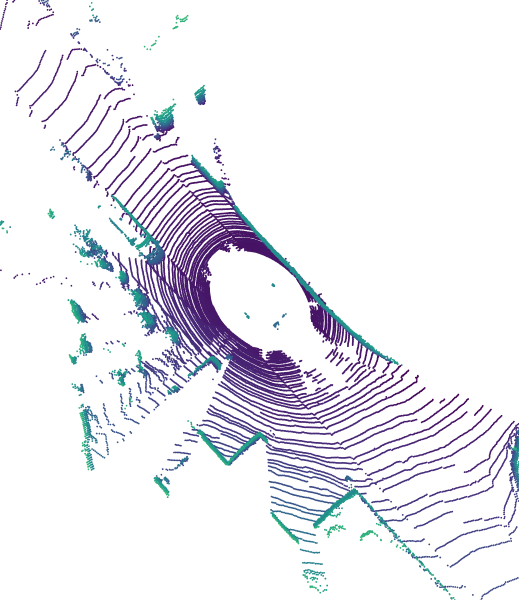} \\
        {\footnotesize Real scene - I}   & \footnotesize Generated scene & \footnotesize Real scene - II \\
    \end{tabular}
    \caption{\textbf{Test-time conditioning with 3D representations for scene mixing.} We condition on stitched features from two real scenes, mixing the left half of one scene with the right half of the other. Generated scene (middle) mixes layouts from real scenes: the left part (purple) resembles the first real scene (left), while the right part (green) resembles the second real scene (right). }
    \label{fig:scene_edit}
    \vspace{-0.5cm}
\end{figure}

We propose two applications of this representation conditioning: object in-painting (using diffusion inversion \cite{song2021denoising}) and scene mixing, where features from multiple real scenes are merged to produce a smooth combination (see Figure~\ref{fig:scene_edit}).
For object in-painting, we transport a real scene toward the Gaussian distribution using the ordinary differential equations defined by the SiT velocity field. 
Noise is added to the tokens within a user-defined mask, after which the process is reversed: denoising is applied starting from this noise while supervising the internal representations of the masked tokens with 3D features of the target object category. The object features are extracted via averaging over the target class on a single scene. The result is a reconstructed scene in which the masked region is filled with points that are both consistent with the surroundings and aligned with the conditioned object. Figure~\ref{fig:inpaint} presents qualitative examples of inpainting.
We note that, these two applications are presented as illustrative examples of the potential of our method, rather than as claims of state-of-the-art performance.

\section{Conclusion}
In this work, we introduced \ours, the first unconditional LiDAR point cloud generation method that integrates RGB image pretrained priors and self-supervised 3D representations. Our proposed training strategy transfers knowledge from large-scale image flow matching models, compensating for the limited scale of LiDAR datasets, while representation alignment with self-supervised 3D features further improves generation quality.  Beyond synthesis, our feature-guided training enables controllable editing at inference, where we present first promising results for object inpainting and scene mixing. We believe this axis will inspire further research. Another promising next step involves leveraging these generations for downstream tasks such as LiDAR segmentation. Experiments on KITTI-360 show that \ours sets a new state of the art, opening new directions for bridging 2D and 3D representation learning in LiDAR generative modeling.
\section*{ACKNOWLEDGMENT}
We acknowledge EuroHPC Joint Undertaking for awarding the project ID EHPC-REG-2024R02-234 access to Karolina, Czech Republic.

{
    \small
    \bibliographystyle{IEEEtran}
    \bibliography{main}

\begin{thebibliography}{10}
\providecommand{\url}[1]{#1}
\csname url@rmstyle\endcsname
\providecommand{\newblock}{\relax}
\providecommand{\bibinfo}[2]{#2}
\providecommand\BIBentrySTDinterwordspacing{\spaceskip=0pt\relax}
\providecommand\BIBentryALTinterwordstretchfactor{4}
\providecommand\BIBentryALTinterwordspacing{\spaceskip=\fontdimen2\font plus
\BIBentryALTinterwordstretchfactor\fontdimen3\font minus \fontdimen4\font\relax}
\providecommand\BIBforeignlanguage[2]{{%
\expandafter\ifx\csname l@#1\endcsname\relax
\typeout{** WARNING: IEEEtran.bst: No hyphenation pattern has been}%
\typeout{** loaded for the language `#1'. Using the pattern for}%
\typeout{** the default language instead.}%
\else
\language=\csname l@#1\endcsname
\fi
#2}}

\bibitem{gschwandtner2011blensor}
M.~Gschwandtner, R.~Kwitt, A.~Uhl, and W.~Pree, ``{BlenSor}: Blender sensor simulation toolbox,'' in \emph{ISVC}, 2011.

\bibitem{fang2020augmentedLiDAR}
J.~Fang, D.~Zhou, F.~Yan, T.~Zhao, F.~Zhang, Y.~Ma, L.~Wang, and R.~Yang, ``Augmented lidar simulator for autonomous driving,'' in \emph{ARL}, 2020.

\bibitem{fang2021LiDARaug}
J.~Fang, X.~Zuo, D.~Zhou, S.~Jin, S.~Wang, and L.~Zhang, ``{LiDAR-Aug}: A general rendering-based augmentation framework for {3D} object detection,'' in \emph{CVPR}, 2021.

\bibitem{yue2018LiDARpcgen}
X.~Yue, B.~Wu, S.~A. Seshia, K.~Keutzer, and A.~L. Sangiovanni-Vincentelli, ``A lidar point cloud generator: from a virtual world to autonomous driving,'' in \emph{ICMR}, 2018.

\bibitem{zyrianov2022learning}
V.~Zyrianov, X.~Zhu, and S.~Wang, ``Learning to generate realistic lidar point clouds,'' in \emph{ECCV}, 2022.

\bibitem{xiong_ultralidar_2023}
Y.~Xiong, W.-C. Ma, J.~Wang, and R.~Urtasun, ``Learning compact representations for lidar completion and generation,'' in \emph{CVPR}, 2023.

\bibitem{song2019generative}
Y.~Song and S.~Ermon, ``Generative modeling by estimating gradients of the data distribution,'' in \emph{NeurIPS}, 2019.

\bibitem{ho_denoising_2020}
J.~Ho, A.~Jain, and P.~Abbeel, ``Denoising {Diffusion} {Probabilistic} {Models},'' in \emph{NeurIPS}, 2020.

\bibitem{song_score-based_2020}
Y.~Song, J.~Sohl-Dickstein, D.~P. Kingma, A.~Kumar, S.~Ermon, and B.~Poole, ``Score-based generative modeling through stochastic differential equations,'' in \emph{ICLR}, 2021.

\bibitem{rombach_high-resolution_2022}
R.~Rombach, A.~Blattmann, D.~Lorenz, P.~Esser, and B.~Ommer, ``High-{Resolution} {Image} {Synthesis} with {Latent} {Diffusion} {Models},'' in \emph{CVPR}, 2022.

\bibitem{podell2024sdxl}
D.~Podell, Z.~English, K.~Lacey, A.~Blattmann, T.~Dockhorn, J.~M{\"u}ller, J.~Penna, and R.~Rombach, ``{SDXL}: Improving latent diffusion models for high-resolution image synthesis,'' in \emph{ICLR}, 2024.

\bibitem{esser2024scaling}
P.~Esser, S.~Kulal, A.~Blattmann, R.~Entezari, J.~M{\"u}ller, H.~Saini, Y.~Levi, D.~Lorenz, A.~Sauer, F.~Boesel, \emph{et~al.}, ``Scaling rectified flow transformers for high-resolution image synthesis,'' in \emph{ICML}, 2024.

\bibitem{yu_representation_2025}
S.~Yu, S.~Kwak, H.~Jang, J.~Jeong, J.~Huang, J.~Shin, and S.~Xie, ``Representation {Alignment} for {Generation}: {Training} {Diffusion} {Transformers} {Is} {Easier} {Than} {You} {Think},'' in \emph{ICLR}, 2025.

\bibitem{leng_repa-e_2025}
X.~Leng, J.~Singh, Y.~Hou, Z.~Xing, S.~Xie, and L.~Zheng, ``{REPA}-{E}: {Unlocking} {VAE} for {End}-to-{End} {Tuning} with {Latent} {Diffusion} {Transformers},'' \emph{arXiv preprint arXiv:2504.10483}, 2025.

\bibitem{tian_u-repa_2025}
Y.~Tian, H.~Chen, M.~Zheng, Y.~Liang, C.~Xu, and Y.~Wang, ``U-{REPA}: {Aligning} {Diffusion} {U}-{Nets} to {ViTs},'' \emph{arXiv:2503.18414}, 2025.

\bibitem{yao_reconstruction_2025}
J.~Yao, B.~Yang, and X.~Wang, ``Reconstruction vs. {Generation}: {Taming} {Optimization} {Dilemma} in {Latent} {Diffusion} {Models},'' in \emph{CVPR}, 2025.

\bibitem{oquab2024dinov2}
M.~Oquab, T.~Darcet, T.~Moutakanni, H.~Vo, M.~Szafraniec, V.~Khalidov, P.~Fernandez, D.~Haziza, F.~Massa, A.~El-Nouby, \emph{et~al.}, ``{DINOv2: Learning Robust Visual Features without Supervision},'' \emph{{Transactions on Machine Learning Research Journal}}, pp. 1--31, Jan. 2024.

\bibitem{nunes2024scaling}
L.~Nunes, R.~Marcuzzi, B.~Mersch, J.~Behley, and C.~Stachniss, ``Scaling diffusion models to real-world 3d lidar scene completion,'' in \emph{CVPR}, 2024.

\bibitem{martyniuk2025lidpm}
T.~Martyniuk, G.~Puy, A.~Boulch, R.~Marlet, and R.~de~Charette, ``Lidpm: Rethinking point diffusion for lidar scene completion,'' in \emph{IV}, 2025.

\bibitem{nakashima_fast_2025}
K.~Nakashima, X.~Liu, T.~Miyawaki, Y.~Iwashita, and R.~Kurazume, ``Fast lidar data generation with rectified flows,'' in \emph{ICRA}, 2025.

\bibitem{nakashima_lidar_2024}
K.~Nakashima and R.~Kurazume, ``{LiDAR} {Data} {Synthesis} with {Denoising} {Diffusion} {Probabilistic} {Models},'' in \emph{ICRA}, 2024.

\bibitem{hu_rangeldm_2024}
Q.~Hu, Z.~Zhang, and W.~Hu, ``{RangeLDM}: {Fast} {Realistic} {LiDAR} {Point} {Cloud} {Generation},'' in \emph{ECCV}, 2024.

\bibitem{ran_towards_2024}
H.~Ran, V.~Guizilini, and Y.~Wang, ``Towards {Realistic} {Scene} {Generation} with {LiDAR} {Diffusion} {Models},'' in \emph{CVPR}, 2024.

\bibitem{caesar2020nuscenes}
H.~Caesar, V.~Bankiti, A.~H. Lang, S.~Vora, V.~E. Liong, Q.~Xu, A.~Krishnan, Y.~Pan, G.~Baldan, and O.~Beijbom, ``nuscenes: A multimodal dataset for autonomous driving,'' in \emph{CVPR}, 2020.

\bibitem{liao2022kitti}
Y.~Liao, J.~Xie, and A.~Geiger, ``Kitti-360: A novel dataset and benchmarks for urban scene understanding in 2d and 3d,'' \emph{IEEE TPAMI}, vol.~45, no.~3, pp. 3292--3310, 2022.

\bibitem{deng2009imagenet}
J.~Deng, W.~Dong, R.~Socher, L.-J. Li, K.~Li, and L.~Fei-Fei, ``Imagenet: A large-scale hierarchical image database,'' in \emph{CVPR}, 2009.

\bibitem{schuhmann2022laion}
C.~Schuhmann, R.~Beaumont, R.~Vencu, C.~Gordon, R.~Wightman, M.~Cherti, T.~Coombes, A.~Katta, C.~Mullis, M.~Wortsman, \emph{et~al.}, ``Laion-5b: An open large-scale dataset for training next generation image-text models,'' \emph{NeurIPS}, 2022.

\bibitem{yan2025olidm}
T.~Yan, J.~Yin, X.~Lang, R.~Yang, C.-Z. Xu, and J.~Shen, ``Olidm: Object-aware lidar diffusion models for autonomous driving,'' in \emph{AAAI}, 2025.

\bibitem{kirby2024logen}
E.~Kirby, M.~Chen, R.~Marlet, and N.~Samet, ``Logen: Toward lidar object generation by point diffusion,'' \emph{BMVC}, 2025.

\bibitem{milioto_rangenet_2019}
A.~Milioto, I.~Vizzo, J.~Behley, and C.~Stachniss, ``{RangeNet} ++: {Fast} and {Accurate} {LiDAR} {Semantic} {Segmentation},'' in \emph{IROS}, 2019.

\bibitem{ando_rangevit_2023}
A.~Ando, S.~Gidaris, A.~Bursuc, G.~Puy, A.~Boulch, and R.~Marlet, ``Rangevit: Towards vision transformers for 3d semantic segmentation in autonomous driving,'' in \emph{CVPR}, 2023.

\bibitem{buburuzan_mobi_2025}
A.~Buburuzan, A.~Sharma, J.~Redford, P.~K. Dokania, and R.~Mueller, ``Mobi: Multimodal object inpainting using diffusion models,'' in \emph{CVPR}, 2025.

\bibitem{zhu2025spiral}
D.~Zhu, Y.~Hu, Y.~Liu, D.~Lu, L.~Kong, and S.~Ilic, ``Spiral: Semantic-aware progressive lidar scene generation and understanding,'' in \emph{NeurIPS}, 2025.

\bibitem{triess_scan-based_2020}
L.~T. Triess, D.~Peter, C.~B. Rist, and J.~M. Zöllner, ``Scan-based {Semantic} {Segmentation} of {LiDAR} {Point} {Clouds}: {An} {Experimental} {Study},'' in \emph{IV}, 2020.

\bibitem{goodfellow2014generative}
I.~Goodfellow, J.~Pouget-Abadie, M.~Mirza, B.~Xu, D.~Warde-Farley, S.~Ozair, A.~Courville, and Y.~Bengio, ``Generative adversarial networks,'' \emph{NeurIPS}, 2014.

\bibitem{van2016conditional}
A.~Van~den Oord, N.~Kalchbrenner, L.~Espeholt, O.~Vinyals, A.~Graves, \emph{et~al.}, ``Conditional image generation with pixelcnn decoders,'' \emph{NeurIPS}, 2016.

\bibitem{lipmanflow}
Y.~Lipman, R.~T. Chen, H.~Ben-Hamu, M.~Nickel, and M.~Le, ``Flow matching for generative modeling,'' in \emph{ICLR}, 2023.

\bibitem{brock2018large}
A.~Brock, J.~Donahue, and K.~Simonyan, ``Large scale {GAN} training for high fidelity natural image synthesis,'' in \emph{ICLR}, 2019.

\bibitem{wang_repa_2025}
Z.~Wang, W.~Zhao, Y.~Zhou, Z.~Li, Z.~Liang, M.~Shi, X.~Zhao, P.~Zhou, K.~Zhang, Z.~Wang, K.~Wang, and Y.~You, ``{REPA} {Works} {Until} {It} {Doesn}'t: {Early}-{Stopped}, {Holistic} {Alignment} {Supercharges} {Diffusion} {Training},'' \emph{arXiv preprint arXiv:2505.16792}, 2025.

\bibitem{wu2025sonata}
X.~Wu, D.~DeTone, D.~Frost, T.~Shen, C.~Xie, N.~Yang, J.~Engel, R.~Newcombe, H.~Zhao, and J.~Straub, ``Sonata: Self-supervised learning of reliable point representations,'' in \emph{CVPR}, 2025.

\bibitem{puy2024three}
G.~Puy, S.~Gidaris, A.~Boulch, O.~Sim{\'e}oni, C.~Sautier, P.~P{\'e}rez, A.~Bursuc, and R.~Marlet, ``Three pillars improving vision foundation model distillation for lidar,'' in \emph{CVPR}, 2024.

\bibitem{peng2023openscene}
S.~Peng, K.~Genova, C.~M. Jiang, A.~Tagliasacchi, M.~Pollefeys, and T.~Funkhouser, ``{OpenScene}: {3D} scene understanding with open vocabularies,'' in \emph{CVPR}, 2023.

\bibitem{ma_sit_2024}
N.~Ma, M.~Goldstein, M.~S. Albergo, N.~M. Boffi, E.~Vanden-Eijnden, and S.~Xie, ``Sit: Exploring flow and diffusion-based generative models with scalable interpolant transformers,'' in \emph{ECCV}, 2024.

\bibitem{albergo2023stochastic}
M.~S. Albergo, N.~M. Boffi, and E.~Vanden-Eijnden, ``Stochastic interpolants: A unifying framework for flows and diffusions,'' \emph{arXiv preprint arXiv:2303.08797}, 2023.

\bibitem{behley2019iccv}
J.~Behley, M.~Garbade, A.~Milioto, J.~Quenzel, S.~Behnke, C.~Stachniss, and J.~Gall, ``{SemanticKITTI: A Dataset for Semantic Scene Understanding of LiDAR Sequences},'' in \emph{ICCV}, 2019.

\bibitem{choy2019minknet}
C.~Choy, J.~Gwak, and S.~Savarese, ``{4D} spatio-temporal {ConvNets}: {Minkowski} convolutional neural networks,'' in \emph{CVPR}, 2019.

\bibitem{spvnas}
H.~Tang, Z.~Liu, S.~Zhao, Y.~Lin, J.~Lin, H.~Wang, and S.~Han, ``Searching efficient 3d architectures with sparse point-voxel convolution,'' in \emph{ECCV}, 2020.

\bibitem{puligilla_lip-loc_2024}
S.~S. Puligilla, M.~Omama, H.~Zaidi, U.~S. Parihar, and M.~Krishna, ``{LIP}-{Loc}: {LiDAR} {Image} {Pretraining} for {Cross}-{Modal} {Localization},'' in \emph{WACVW}, 2024.

\bibitem{song2021denoising}
J.~Song, C.~Meng, and S.~Ermon, ``Denoising diffusion implicit models,'' in \emph{ICLR}, 2021.

\end{thebibliography}
}

\end{document}